\documentclass[9pt,twocolumn,twoside]{osajnl}

\usepackage{times}
\usepackage{epsfig}
\usepackage{graphicx}
\usepackage{float}
\usepackage{wrapfig}

\usepackage{amsmath,amssymb}

\usepackage{bm,xspace}
\usepackage{comment}
\usepackage{verbatim}
\usepackage{multirow}
\usepackage{balance}
\usepackage{url}
\usepackage{booktabs}
\usepackage{etoolbox,siunitx}
\usepackage{calc}
\usepackage{pifont,hologo}
\usepackage{color}

\usepackage[super]{nth}
\usepackage{nicefrac}
\def\aa{\mathbf{a}}

\def\ss{\mathbf{s}}
\def\tt{\mathbf{t}}

\def\AA{\mathbf{A}}

\DeclareMathOperator*{\argmax}{arg\,max}

\DeclareMathSymbol{@}{\mathord}{letters}{"3B}

\def\latex/{\LaTeX}
\def\bibtex/{\hologo{BibTeX}}

\usepackage[official]{eurosym}

\newcommand{\hide}[1]{}

\usepackage{amsmath} 
\usepackage{amssymb}
\usepackage{bm}
\usepackage{acro}
\usepackage{balance}
\usepackage{booktabs}
\usepackage{pifont}
\usepackage{color}
\usepackage[]{graphicx} 
\usepackage{hyperref}
\usepackage{siunitx}    
\usepackage{times}
\usepackage{tabularx} 
\usepackage{subcaption}
\usepackage{xcolor}
\usepackage{enumitem}
\usepackage{adjustbox}
\usepackage{tabularx,array}
\usepackage{pdflscape}
\usepackage{rotating}
\usepackage{tabularx}
\usepackage{wrapfig}

\usepackage[TS1,T1]{fontenc}
\DeclareSIUnit[number-unit-product = {}]{\inch}{\textquotedbl}

\usepackage{tikz,pgfplots}
\usetikzlibrary{matrix}
\pgfplotsset{compat=1.15}
\usepackage{etex} 
\usetikzlibrary{arrows,automata}
\usetikzlibrary{positioning}
\usepgfplotslibrary{groupplots}
\usepackage{tikz-3dplot}
\tikzset{
	state/.style={
		rectangle,
		rounded corners,
		draw=black, very thick,
		minimum height=2em,
		inner sep=2pt,
		text centered,
	},
}
\tikzset{
	info/.style={
		rectangle,
		draw=black, thin,
		minimum height=2em,
		inner sep=2pt,
		text centered,
	},
}
\usetikzlibrary{positioning}
\usetikzlibrary{calc}
\usetikzlibrary{arrows,shapes,backgrounds}
\usepackage{nameref} 

\newcolumntype{Y}{>{\centering\arraybackslash}X}

\newcommand{\change}[1]{#1}

\newcommand{\rebuttal}[1]{#1}

\newcommand{\final}[1]{{\color{black}#1}}

\newlist{todolist}{itemize}{2}
\setlist[todolist]{label=$\square$}

\newcolumntype{C}[1]{>{\centering\let\newline\\\arraybackslash\hspace{0pt}}m{#1}}

\DeclareAcronym{TWR}{
    short = thrust-to-weight,
    long = thrust-to-weight ratio
}

\DeclareAcronym{TIR}{
    short = torque-to-inertia,
    long = torque-to-inertia ratio
}

\DeclareAcronym{SBC}{
    short = SBC,
    long = single-board computer
}

\DeclareAcronym{ROS}{
    short = ROS,
    long = Robot Operating System
}

\DeclareAcronym{FPV}{
    short = FPV,
    long = first-person-view
}

\DeclareAcronym{HIL}{
    short = HIL,
    long = hardware-in-the-loop
}

\DeclareAcronym{MAV}{
    short = MAV,
    long = micro-aerial vehicle
}

\DeclareAcronym{ASIC}{
    short = ASIC,
    long = application-specific integrated circuit
}

\DeclareAcronym{MPC}{
    short = MPC,
    long = model-predictive control
}

\DeclareAcronym{ISP}{
    short = ISP,
    long = image-signal processor
}

\DeclareAcronym{VIO}{
    short = VIO,
    long = visual-inertial odometry
}

\DeclareAcronym{IMU}{
    short = IMU,
    long = inertial measurement unit
}

\DeclareAcronym{INDI}{
    short = INDI,
    long = incremental non-linear dynamic inversion
}

\DeclareAcronym{RMSE}{
    short = RMSE,
    long = root-mean-square error
}

\DeclareAcronym{ATE}{
    short = ATE,
    long = absolute tracking error
}

\DeclareAcronym{BEM}{
    short = BEM,
    long = blade-element momentum theory
}

\tikzstyle{mybox} = [draw=black, fill=gray!20, very thick, rectangle, rounded corners, inner sep=1pt, inner ysep=2pt]
\tikzstyle{fancytitle} =[fill=black, text=white]

\DeclareSIUnit\baud{Baud}
\DeclareSIUnit\flops{FLOPS}
\DeclareSIUnit\g{g}
\DeclareSIUnit\inch{inch}

\newcommand{\bs}{\boldsymbol}

\journal{ol} 

\setboolean{shortarticle}{false}

\title{Reaching the Limit in Autonomous Racing: \\ 
Optimal Control versus Reinforcement Learning
}

\author[$1$, *]{Yunlong Song}
\author[$1$]{Angel Romero}
\author[$2$]{Matthias M\"uller}
\author[$3$]{Vladlen Koltun}
\author[$1$]{Davide Scaramuzza}

\affil[$1$]{Authors are with the Robotics and Perception Group, UZH, Zurich, Switzerland.}
\affil[$2$]{The Author is with Intel, Munich, Germany.}
\affil[$3$]{The Author is with Intel Labs, Jackson, WY, USA.}
\affil[*]{Corresponding author: song@ifi.uzh.ch}

\dates{This is the accepted version of Science Robotics Vol. 8, Issue 82 \\ DOI: 10.1126/scirobotics.adg1462 (2023) }

\begin{abstract}
A central question in robotics is how to design a control system for an agile mobile robot.
This paper studies this question systematically, focusing on a challenging setting: autonomous drone racing. 
We show that a neural network controller trained with reinforcement learning (RL) outperforms optimal control (OC) methods in this setting.
We then investigate which fundamental factors have contributed to the success of RL or have limited OC. 
Our study indicates that the fundamental advantage of RL over OC is not that it
optimizes its objective better but that it optimizes a better objective.
OC decomposes the problem into planning and control with an explicit intermediate representation, such as a trajectory, that serves as an interface.
This decomposition limits the range of behaviors that can be expressed by the controller, leading to inferior control performance when facing unmodeled effects.
In contrast, RL can directly optimize a task-level objective and can leverage domain randomization to cope with model uncertainty, allowing the discovery of more robust control responses.
Our findings allow us to push an agile drone to its maximum performance, achieving a peak acceleration greater than 12 g and a peak velocity of~\SI{108}{\kilo\meter\per\hour}. Our policy achieves superhuman control within minutes of training on a standard workstation. 
This work presents a milestone in agile robotics and sheds light on the role of RL and OC in robot control.
\end{abstract}

\begin{document}

\maketitle

\section*{Multimedia Material}
A video of the experiments can be found at \\
\quad {\small\url{https://youtu.be/HGULBBAo5lA}}.

\section*{Introduction}
\label{sec:introduction}

\begin{figure*}
\centering
\includegraphics[width=1\textwidth]{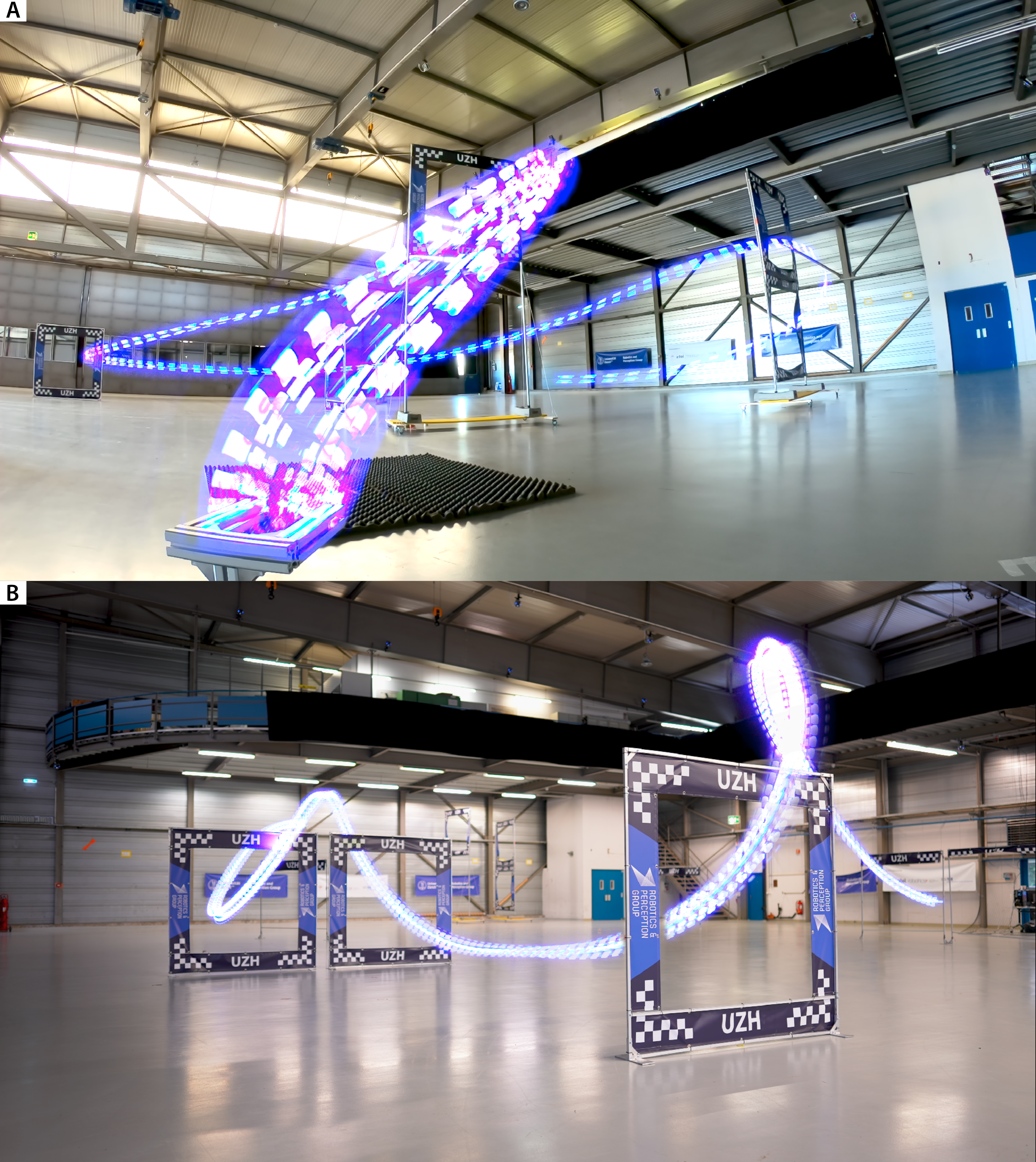}
\caption{
\textbf{Time-lapse illustrations of a high-performance racing drone controlled by our RL policy.}
We test on two race tracks, the Split-S track \textbf{(A)} and the Marv track \textbf{(B)}; both are designed by world-class human drone pilots.   
Our RL policy achieves a peak speed of \SI{108}{\kilo\meter\per\hour} and a peak acceleration greater than \SI{12}{\g} in an indoor flying arena.
Recordings of the experiments can be found in Movie~1.
} 
\label{fig: overview}
\end{figure*}

The design of control systems for agile mobile robots is one of the central challenges in robotics.
Control systems are at the core of every real-world robot and are deployed in an ever-increasing number of applications, such as self-driving cars~\cite{paden2016survey,maurer2016autonomous}, search and rescue~\cite{delmerico2019current}, flying cars~\cite{air_taxi_2020}, autonomous vehicle racing~\cite{Spielberg2019ScienceRobotics, kaufmann2023champion}, navigation of stratospheric balloons~\cite{Bellemare2020Nature}, and extraterrestrial exploration~\cite{tzanetos2022ingenuity, grip2022flying}.
\rebuttal{
Two schools of thought exist for the development of such systems: model-based optimal control~(OC) and learning-centric reinforcement learning~(RL); both have achieved impressive results. }

\rebuttal{ OC~\cite{arthur1975applied, bertsekas2012dynamic} relies on the explicit use of an accurate mathematical model within an optimization framework to find an optimal control law for a given dynamical system.
This technique involves solving an optimization problem online to find the control inputs that minimize a cost function while satisfying system constraints. 
OC techniques make use of convex optimization methods such that the problem is solved in real time at high frequency. 
Examples of OC methods are the Linear Quadratic Regulator~(LQR) and its nonlinear finite-horizon counterpart, Model Predictive Control (MPC).}
\rebuttal{
RL~\cite{sutton2018reinforcement}, on the other hand, is a subfield of machine learning that trains an agent to maximize a reward signal in an environment. The agent learns by trial and error, exploring the environment and receiving feedback in the form of rewards or penalties for its actions. RL algorithms can be categorized into three general approaches: value-based, policy-based, and actor-critic.
}

Connections between OC and RL are strong: both seek to find the optimal mapping from observations to control commands, and both are rooted in the principle of optimality derived from dynamic programming~\cite{bertsekas2019reinforcement}. 
Yet there is a gap: RL can handle problems of high dimensionality and long planning horizons, albeit mostly in simulation~\cite{mnih2015human, silver2016mastering, vinyals2019grandmaster, Bellemare2020Nature, degrave2022magnetic}, and has only recently~\cite{kaufmann2023champion} been shown to outperform humans in a real-world sport.
OC, on the other hand, has been successful primarily in tackling control problems in fairly well-understood dynamical systems~\cite{kuindersma2016optimization, bledt2018cheetah, neunert2018whole, falanga2018pampc, neunert2016fast}.  
\rebuttal{
In addition, OC typically assumes that the system being controlled is deterministic, \final{whereas} RL can intrinsically handle both deterministic and stochastic systems.}

In most OC setups~\cite{Romero2021arxiv, bledt2018cheetah, neunert2018whole, falanga2018pampc, neunert2016fast, wieber2016modeling, bjelonic2022offline, kuindersma2016optimization}, the high-level task is first converted into a reference trajectory (planning) and then tracked by a controller (control).
This decomposition of the problem into these distinct layers is greatly favored by the OC methodology, largely due to the interpretability of each component's output and the simplification of the pipeline for real-time control.
The optimization objective used by the control layer is shaped to achieve accurate trajectory tracking or path following and is decoupled from (and usually unrelated to) the high-level task objective. 
As a result, the hierarchical separation of the information between two components leads to systems that can become erratic in the presence of unmodeled dynamics.
In practice, a series of conservative assumptions or approximations are required to counteract model mismatches and maintain controllability, resulting in systems that are no longer optimal. 

\rebuttal{
RL has arisen as an attractive alternative to conventional controller design, demonstrating exceptional performance in various domains such as quadrupedal locomotion over challenging terrain~\cite{LeeSR2020, Miki2022ScienceRobotics}. 
Unlike optimal control, RL uses sampled data to optimize a controller and can manage nonconvex and even sparse objectives, providing \final{substantial} flexibility in the controller design. 
RL has several advantages over model-based optimal control. First, it learns a control policy via offline optimization, enabling the trained policy to efficiently compute control commands during deployment. 
Unlike offline trajectory optimization which generates predetermined trajectories, offline policy optimization focuses on learning a feedback controller that allows real-time adaptation given observation changes.} 
Second, RL can directly optimize a task objective, eliminating the need for explicit intermediate representations such as trajectories. Finally, 
RL can leverage domain randomization in simulation, enabling the learning of a policy that is effective in diverse environments.

Some of the most impressive achievements of RL are beyond the reach of existing OC-based systems. 
However, most of these successes are empirical. 
Less attention has been paid to the systematic study of fundamental factors that have led to the success of RL or have limited OC. 
We argue that this question can be investigated along two axes: the optimization method and the optimization objective. 
On one hand, RL and OC can be viewed as two different optimization methods and we can ask which method can achieve a more \change{robust} solution given the same cost function.
On the other hand, given that RL and OC address a given robot control problem by optimizing different objectives, we can ask which optimization objective can lead to more robust task performance. 
\rebuttal{In this context, robust task performance refers to the controller's capability to consistently perform a given task without sacrificing performance, even in the face of uncertainties and disturbances.}

We perform this investigation in a challenging real-world problem that involves a high-performance robotic system: autonomous drone racing.
The task of drone racing is to fly a quadrotor through a sequence of gates in a given order in minimum time.
For maximal performance, this task requires pushing the aircraft to its physical limits of speed and acceleration.
Tolerance for error is low: a small mistake can lead to a catastrophic crash or a strong penalty on lap time. 
Thus, suboptimal control policies readily manifest themselves in reduced task performance, making drone racing a particularly demanding and instructive setting for testing the limits of control design paradigms~\cite{moon2019challenges, de2022sensing, Foehn20rss, de2021artificial}.

Our main contribution is the study of reinforcement learning and optimal control from the fundamental perspective of the \final{optimization method and optimization objective}. 
Our results indicate that RL does not outperform OC because RL optimizes its objective better. Rather, RL outperforms OC because it optimizes a better objective.
%
Specifically, RL directly maximizes a task-level objective, which leads to more \change{robust control performance} in the presence of unmodeled dynamics.
In the drone racing context, RL can optimize a highly nonlinear and nonconvex gate-progress reward directly, removing the need for a reference time trajectory or a continuous 3D path.
%
In contrast, OC is limited by its decomposition of the problem into planning and control, which requires an intermediate representation in the form of a trajectory or path, thus limiting the range of control policies that can be expressed by the system.
\rebuttal{
In addition, RL can leverage domain randomization to achieve extra robustness and avoid overfitting, where the agent is trained on a variety of simulated environments with varying settings.}

Beyond the fundamental study, our work contributes an RL-based controller that delivers the highest performance ever demonstrated on an autonomous racing drone.
%
Using \final{an agile autonomous drone}, our controller achieves a peak acceleration greater than \SI{12}{\g} and a peak velocity of~\SI{108}{\kilo\meter\per\hour}. 
\final{Figure~\ref{fig: overview} displays time-lapse illustrations of the racing drone controlled by our RL policy in an indoor flying arena.} \final{Additionally}, our controller demonstrated superhuman performance, outracing three professional human pilots in a public event.
Notably, our controller is trained purely in simulation, in minutes on a standard workstation, and transferred to the real world zero-shot. 

\section*{Results}  
\label{sec:experiments}

\subsection*{RL versus OC}
\label{sec:rlvscontrol}

\begin{figure*}[!htp]
\begin{tikzpicture}
\node at (0,0) (main) [text width = 1\linewidth] {
\renewcommand*{\arraystretch}{1.5}
    \centering
    \adjustbox{max width=1.0\linewidth}{
        \fontsize{9pt}{15pt}\selectfont     
        \begin{tabular}{c c c c}
           & Trajectory Tracking & Contouring Control & Reinforcement Learning \\
            \toprule
            \bigskip
            \begin{sideways} Nominal Drone Model \end{sideways}    &
            \begin{subfigure}[t]{0.30\textwidth}
                \includegraphics[width=1.0\textwidth]{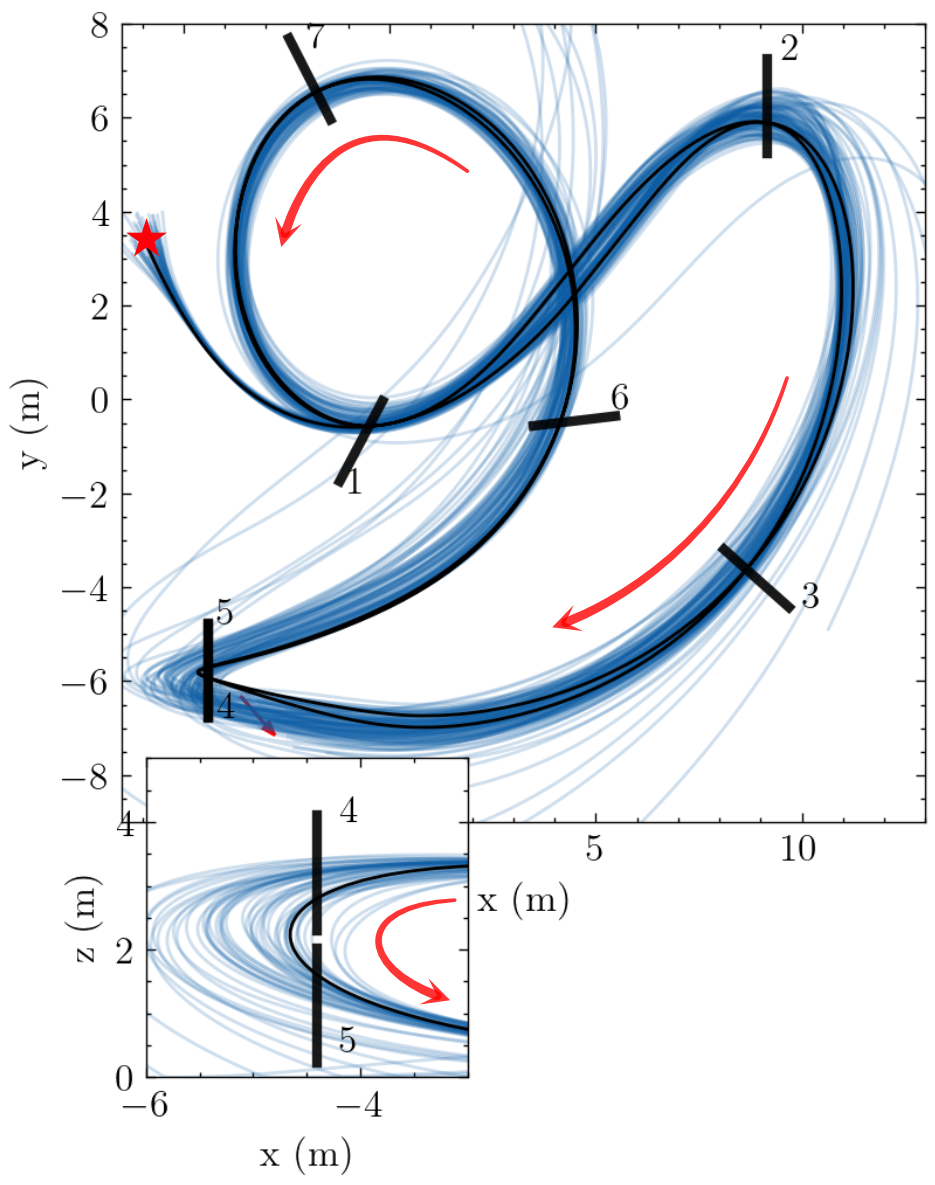}
            \end{subfigure} &
            \begin{subfigure}[t]{0.30\textwidth}
                \includegraphics[width=1.0\textwidth]{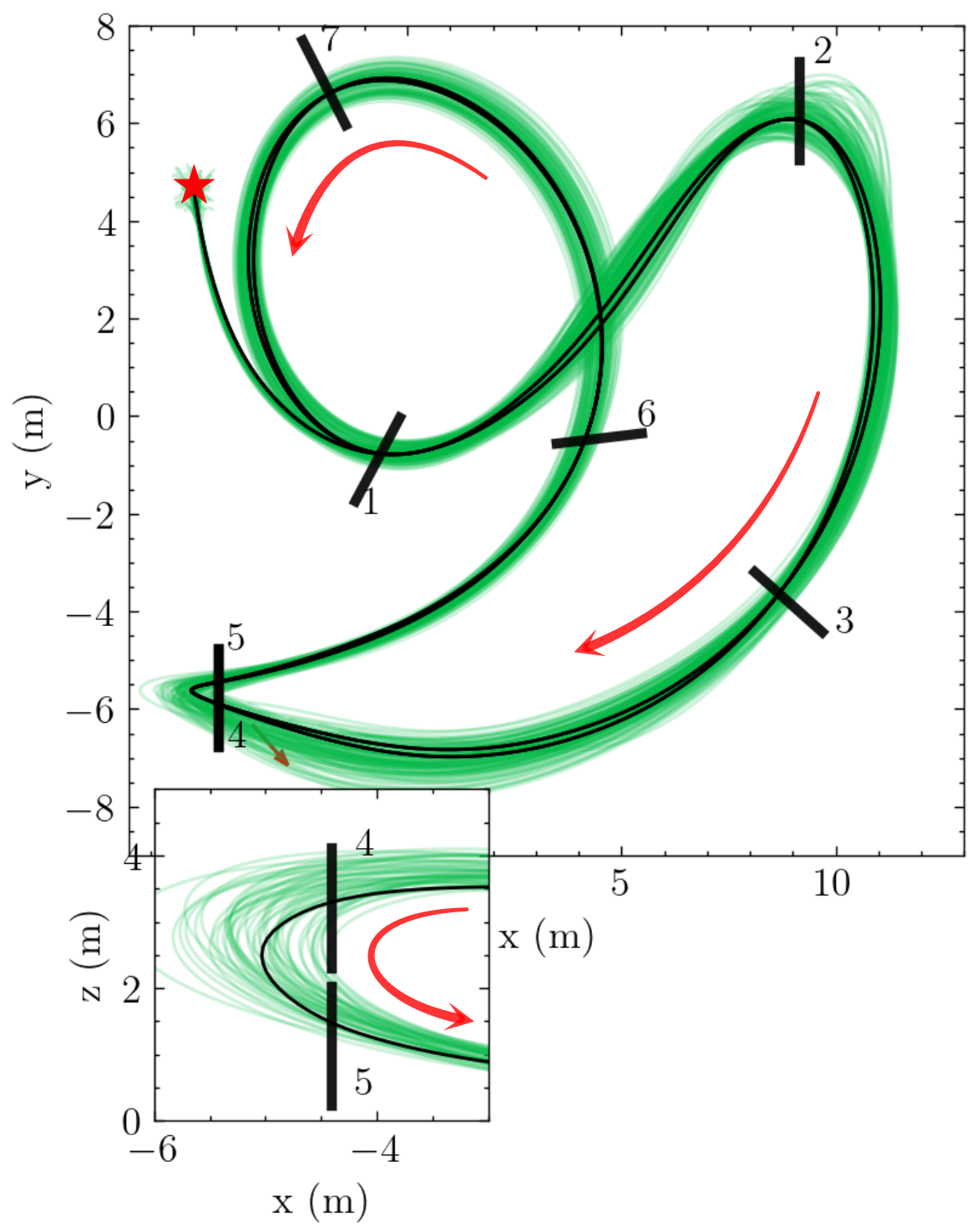}
            \end{subfigure} & 
            \begin{subfigure}[t]{0.30\textwidth}
                \includegraphics[width=1.0\textwidth]{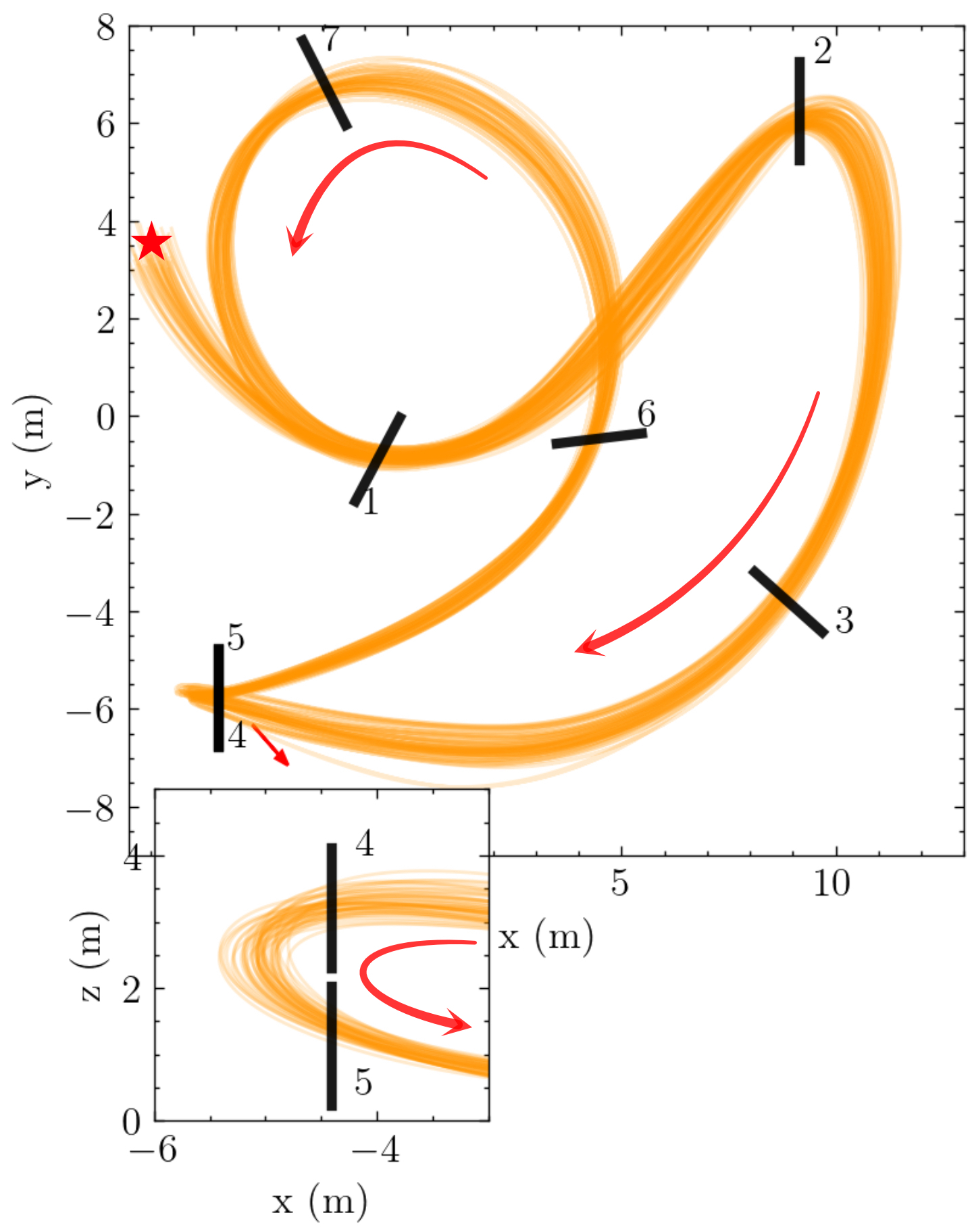}
            \end{subfigure} \\ \toprule
            \bigskip
            \begin{sideways}  Realistic Drone Model  \end{sideways}    &
            \begin{subfigure}[t]{0.30\textwidth}
                \includegraphics[width=1.0\textwidth]{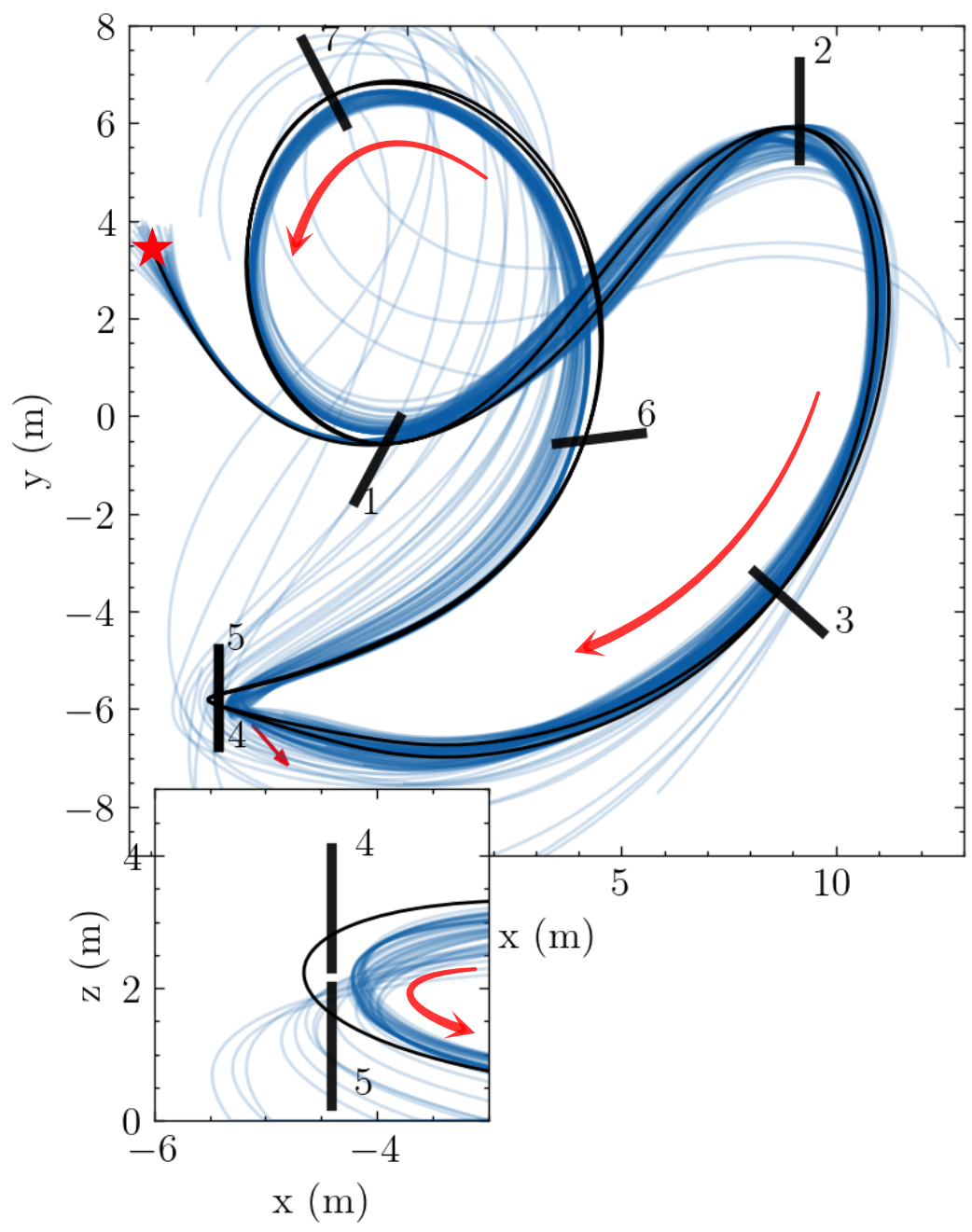}
            \end{subfigure} &
            \begin{subfigure}[t]{0.30\textwidth}
                \includegraphics[width=1.0\textwidth]{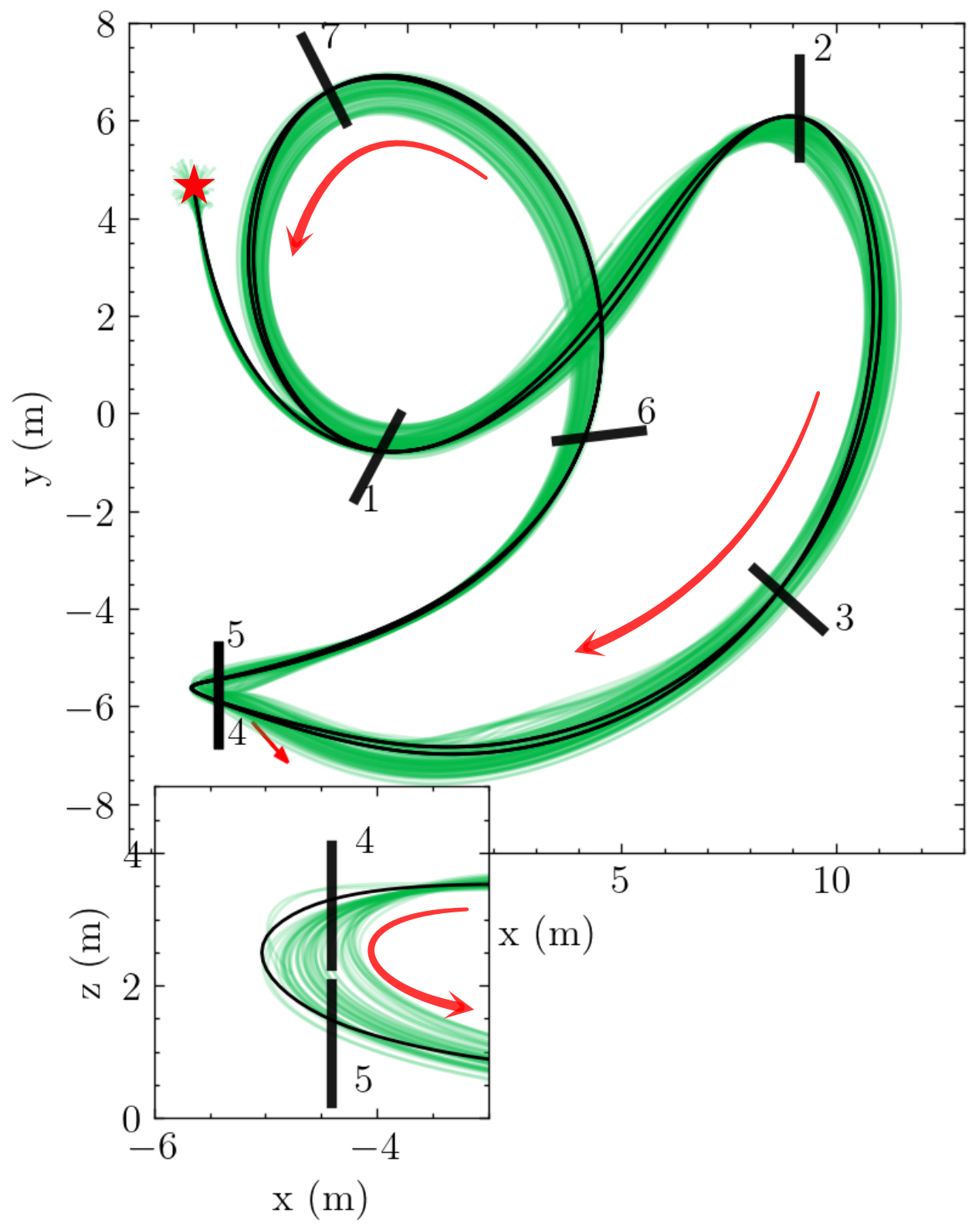}
            \end{subfigure} & 
            \begin{subfigure}[t]{0.30\textwidth}
                \includegraphics[width=1.0\textwidth]{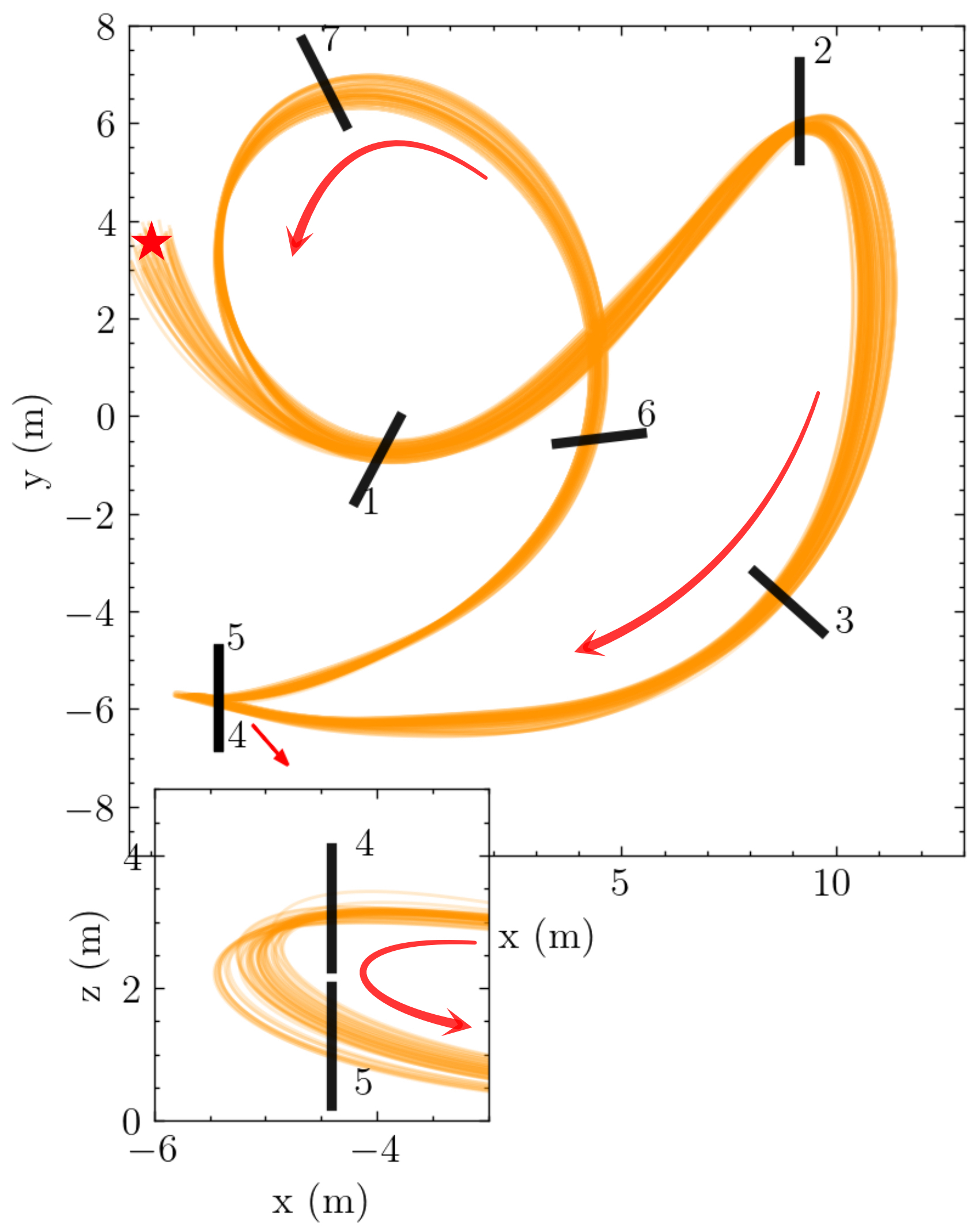}
            \end{subfigure} \\
        \end{tabular}
    }
    \resizebox{\textwidth}{!}{%
        \adjustbox{max width=1.0\linewidth}{
        \begin{tabular}{ c c c | c c | c c }
                \toprule
                \multirow{2}{*}{Drone Model} 
                & \multicolumn{2}{c}{ Trajectory Tracking} 
                & \multicolumn{2}{c}{ Contouring Control} 
                & \multicolumn{2}{c}{ Reinforcement Learning } \\ \cline{2-7}
                & Lap Time~[\SI{}{\second}] & Success Rate~[\%] 
                & Lap Time~[\SI{}{\second}] & Success Rate~[\%] 
                & Lap Time~[\SI{}{\second}] & Success Rate~[\%] \\
                \hline
                 Nominal
                 & \textbf{4.92$\pm$ 0.10}  & 44.0 & 5.03 $\pm$ 0.18 & 76.0 & 5.14  $\pm$ 0.09 & \textbf{100.0} \\ \hline
                 Realistic  
                 & \textbf{--} & 0.0  & 5.34 $\pm$ 0.27  & 20.0  & \textbf{5.26 $\pm$ 0.32}  & \textbf{100.0}   \\ \hline
                \textbf{Real World} 
                 & \textbf{--} & 0.0 &  5.54 $\pm$ 0.21  & 50.0  & \textbf{5.35 $\pm$ 0.15}  & \textbf{85.0} \\
                 \bottomrule
            \end{tabular} 
        }
    }
    };
\node at ([xshift=0.2cm, yshift=-0.55cm]main.north west) [anchor = north west, inner sep = 0pt, outer sep = 0pt, text width = 0.5cm, align = left] {
      \includegraphics[]{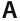}%
      };
\node at ([xshift=0.2cm, yshift=-7.2cm]main.north west) [anchor = north west, inner sep = 0pt, outer sep = 0pt, text width = 0.5cm, align = left] {
      \includegraphics[]{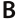}%
      };
\node at ([xshift=0.2cm, yshift=-13.9cm]main.north west) [anchor = north west, inner sep = 0pt, outer sep = 0pt, text width = 0.5cm, align = left] {
      \includegraphics[]{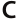}%
      };
\end{tikzpicture}
    \caption{
    \textbf{RL versus OC in autonomous drone racing.}
    Both OC methods are sensitive to initial conditions and unmodeled effects, resulting in performance that is no longer optimal in the presence of unmodeled dynamics (realistic drone model). 
    \textbf{(A)} Trajectories flown with the nominal drone model. 
    \textbf{(B)} Trajectories flown with the realistic drone model. 
    \textbf{(C)} A comparison of the lap time and success rates. \rebuttal{The red arrows indicates the flight direction and the \final{star} symbol shows the starting position. } 
    }
    \label{fig: result1}
\end{figure*}

We compare the RL policy with two state-of-the-art OC methods, which were selected due to their superior control performance in recent work.
The first, referred to as Trajectory Tracking~\cite{Foehn2021science}, relies on offline time-optimal trajectory planning using nonlinear optimization and online tracking using model predictive control~(MPC).
Due to the complexity of the minimum-time problem when considering the full quadrotor dynamics, the time-optimal planning requires hours of computation and thus can only be solved offline. 
The second OC method, referred to as Contouring Control~\cite{Romero2021arxiv}, solves the time-optimal flight problem online by simultaneously maximizing the progress along a reference path and minimizing the vehicle's deviation from the path. 
This nominal path is a continuously differentiable 3D trajectory, which can be generated efficiently using an approximated point-mass model. 
Contouring control leverages efficient path planning and fast trajectory optimization online in a receding horizon fashion. 

We use three different drone models: a nominal drone model based on simple rigid-body dynamics~\cite{yunlong2020flightmare}, a realistic drone model based on blade element momentum theory~\cite{bauersfeld2021neurobem}, and a real-world racing drone~\cite{foehn2022agilicious}. 
All methods use the nominal drone model for optimizing the control systems. 
Namely, both OC methods employ the nominal model in the optimization, and RL uses the nominal model for training.
\change{When training the RL policy, we can incorporate domain randomization by randomizing several parameters, such as the thrust mapping and drag coefficients.}
The realistic drone model and the physical drone are used for testing generalization across different dynamics models in simulation and generalization from simulation to the physical world.
We use the same track as in recent work on high-performance drone racing for benchmark comparisons~\cite{Foehn2021science, Romero2021arxiv}. 
\final{The track was designed by Drone-Racing-League pilot Gabriel Kocher.}

We begin with a large-scale experiment in simulation by testing each approach from 50 different random initial positions. 
We simulate a system delay of \SI{40}{\milli\second} for all methods when testing in simulation. 
Also, when using the nominal model, we additionally randomize the thrust mapping coefficients to simulate unmodeled battery behaviour, such as high voltage drops when flying at very high speeds, and the drag coefficients to simulate unknown aerodynamic effects.
Figure~\ref{fig: result1} shows a visualization of the 50 trajectories flown by each approach using the nominal and realistic models in simulation. We show the top-down view of the entire track and a side view of gates 4 and 5. 
The colored lines are trajectories, and the black lines represent the gates. 
Both OC methods are sensitive to unmodeled dynamics and different initial conditions, \final{whereas} the RL policy maintains high racing performance in all conditions.

\final{Figure~\ref{fig: result1}C} summarizes the quantitative results, including lap time and success rate. 
When testing with the nominal drone model, Trajectory Tracking achieves the best average lap time ($4.92$ s) by tracking a time-optimal trajectory planned with the same nominal model. 
However, it has a low success rate ($44\%$) even with the nominal drone model.
\rebuttal{By definition, a time-optimal trajectory is designed to make the most of the available actuator power at all times. However, due to limited actuation power, control authority may decrease under model mismatches and disturbances, and even slight deviations from the reference state can lead to catastrophic crashes.}
Contouring Control achieves a similar average lap time with the nominal model ($5.03$ s), with a higher success rate ($76\%$). 
RL achieves a lap time of $5.14$ s with the nominal model, with the highest success rate ($100\%$). 
When testing with the realistic drone model, both Trajectory Tracking and Contouring Control fail, with success rates of $0\%$ and $20\%$, respectively.
%
%
The RL policy achieves the best lap time and a $100\%$ success rate, despite the mismatch between the nominal dynamics model experienced during training and the realistic model used at test time.
%
All methods are also tested in the physical world on the physical racing drone. 
Real-world testing introduces additional factors that are not present in the simulation, including aerodynamic effects, variable system delay, and large battery voltage fluctuations. 
As a result, both OC methods fail in this setting. 
Trajectory Tracking crashes the drone immediately after launch, due to unmodeled dynamics that result in a rapid departure from the tracked trajectory while operating at full thrust.
%
\rebuttal{To maintain controllability in the physical world, Foehn et al.~\cite{Foehn2021science} tracked the time-optimal trajectory at a lower thrust bound than what the platform can deliver, yielding a $100\%$ success rate, but at a lap time (\SI{6.12}{\second}) that were far from the optimum.
Contouring Control requires \final{manual} tuning by a human expert, including tuning of parameters that govern the trade-off between path progress maximization and contouring error minimization. These parameters need to be tuned in the physical world at the track. 
Poorly tuned parameters lead to an overly aggressive policy that crashes into gates or an overly conservative one that yields suboptimal lap times. 
As a result, to achieve robust control performance in the real world, Romero et al.~\cite{Romero2021arxiv} had to compromise on speed and achieved an average lap time of~\SI{5.8}{\second}.
In contrast, the RL policy, although trained purely in simulation with the nominal drone model, transfers directly to the physical world, with no fine-tuning, and achieves \final{a better lap} time (average of~\SI{5.35}{\second}). }

\subsection*{Optimization Method versus Optimization Objective}
\label{sec:om-vs-oo}

We have seen that RL outperforms state-of-the-art optimal control methods, achieving higher success rates and lower lap time in the presence of unmodeled dynamics. 
We now ask what is responsible for better racing performance and the robustness of RL.
We study this question along two axes: the optimization method and the optimization objective. 

\subsubsection*{Optimization Method Hypothesis}
The first hypothesis concerns the difference in the optimization method: given the same optimization objective, which method can lead to a better solution? 
RL and OC rely on different optimization techniques to find the best solution for a specific objective. 
Model-free RL optimizes a parameterized policy by following the policy gradients estimated from data generated during the execution of the task. 
\rebuttal{A key challenge in RL is to obtain a good estimator of the policy gradient for policy updates.}
On the other hand, nonlinear OC relies on numerical optimization methods, such as specially structured nonlinear programs (NLP) that can be solved by sequential quadratic programming (SQP) or nonlinear interior point methods. 
The difference in the optimization method \final{influences} the performance of the resulting controller.
\rebuttal{
We thus consider the \final{Optimization Method Hypothesis}: RL outperforms OC because RL, as an optimizer, can achieve better task performance than OC. The difference in the optimization method makes RL more effective than OC.}

To test this hypothesis, we compare RL and nonlinear MPC in optimizing the same objective under the same conditions. 
Specifically, we use RL to follow the time-optimal trajectory by minimizing the same quadratic cost formulation with the same cost matrix as in MPC. 
%
%
The RL policy is trained offline until it is fully converged. 
%
%
%
As shown in Figure~\ref{fig: objective_vs_method}A, the RL policy produces higher average tracking losses than nonlinear MPC when tested with either the nominal drone model or the realistic model.
\rebuttal{Hence, RL cannot find better solutions than OC given the same optimization objective.}
Nevertheless, the RL policy can still fly the vehicle through all gates without collision and achieve the same optimal lap time (\SI{4.9}{\second}) when tested with the nominal model.
When tested with the realistic model, both MPC and RL suffer \final{from performance drops}. MPC yields lower tracking loss on average and a tighter spread across trials.
We conclude that in this condition, when optimizing the same objective, RL does not yield a more effective policy than MPC. 
Also, both MPC and RL yield inferior control performance using the traditional planning-and-control design when facing model mismatch. 
Hence, the explanation for the robust performance of RL that we saw in \final{the previous section} must lie elsewhere.

\subsubsection*{Optimization Objective Hypothesis}
\label{sec:optimization_objective_hypothesis}

\begin{figure*}[h!]
\begin{tikzpicture}
\node at (0,0) (main) [text width = 1.0\linewidth] {
\renewcommand*{\arraystretch}{1.5}
    \centering
    \footnotesize
    \begin{tabular}{p{1\linewidth}}
    \adjustbox{max width=1\linewidth}{
        \begin{tabular}{c c}
            \textbf{Optimization Method}  &\textbf{Optimization Objective}\\
            \begin{subfigure}[t]{0.48\textwidth}
                 \captionsetup{labelformat=empty}
                \includegraphics[width=1\textwidth]{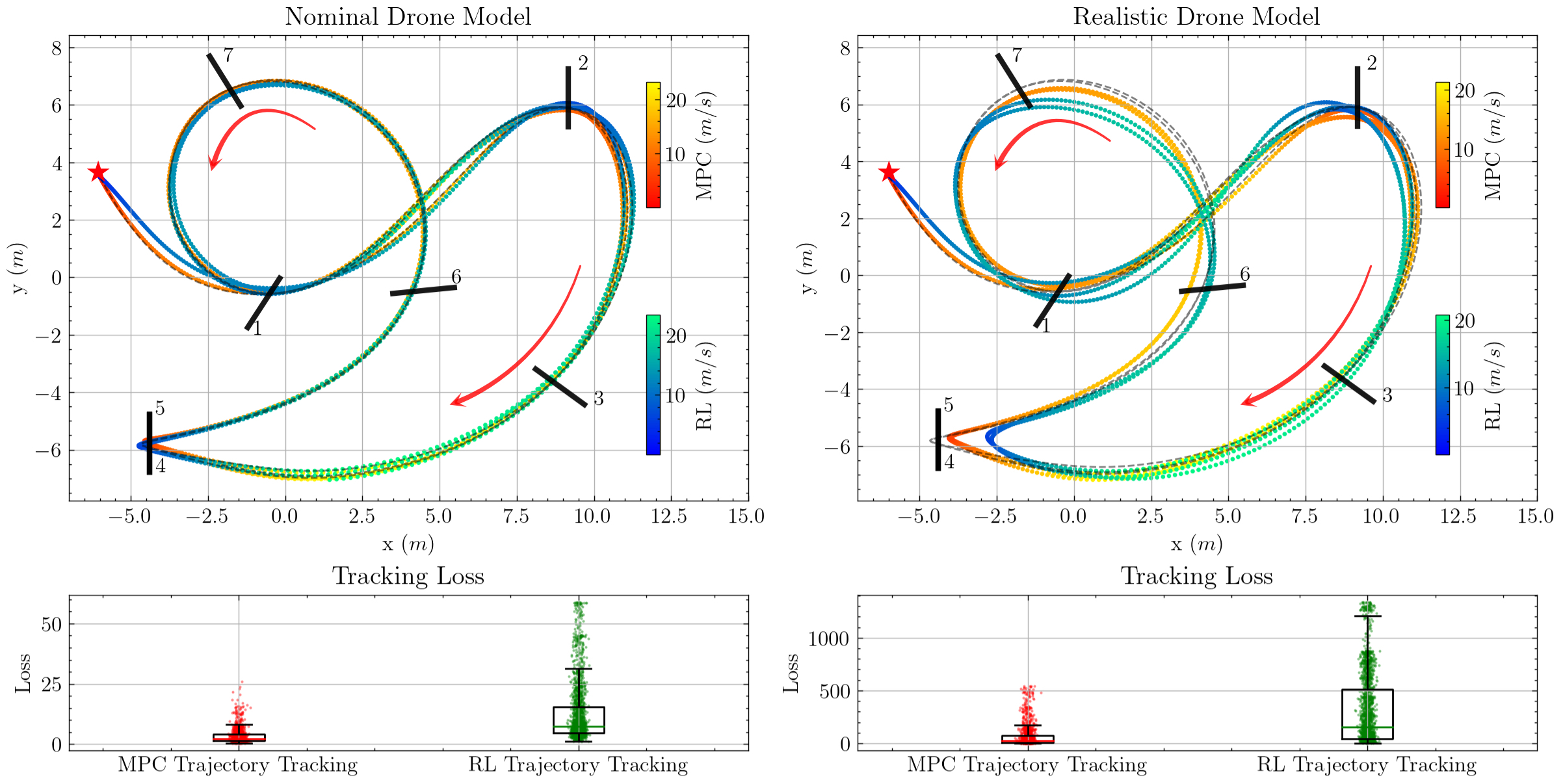} 
            \end{subfigure} &
            \begin{subfigure}[t]{0.48\textwidth}
                 \captionsetup{labelformat=empty}
                \includegraphics[width=0.86\textwidth]{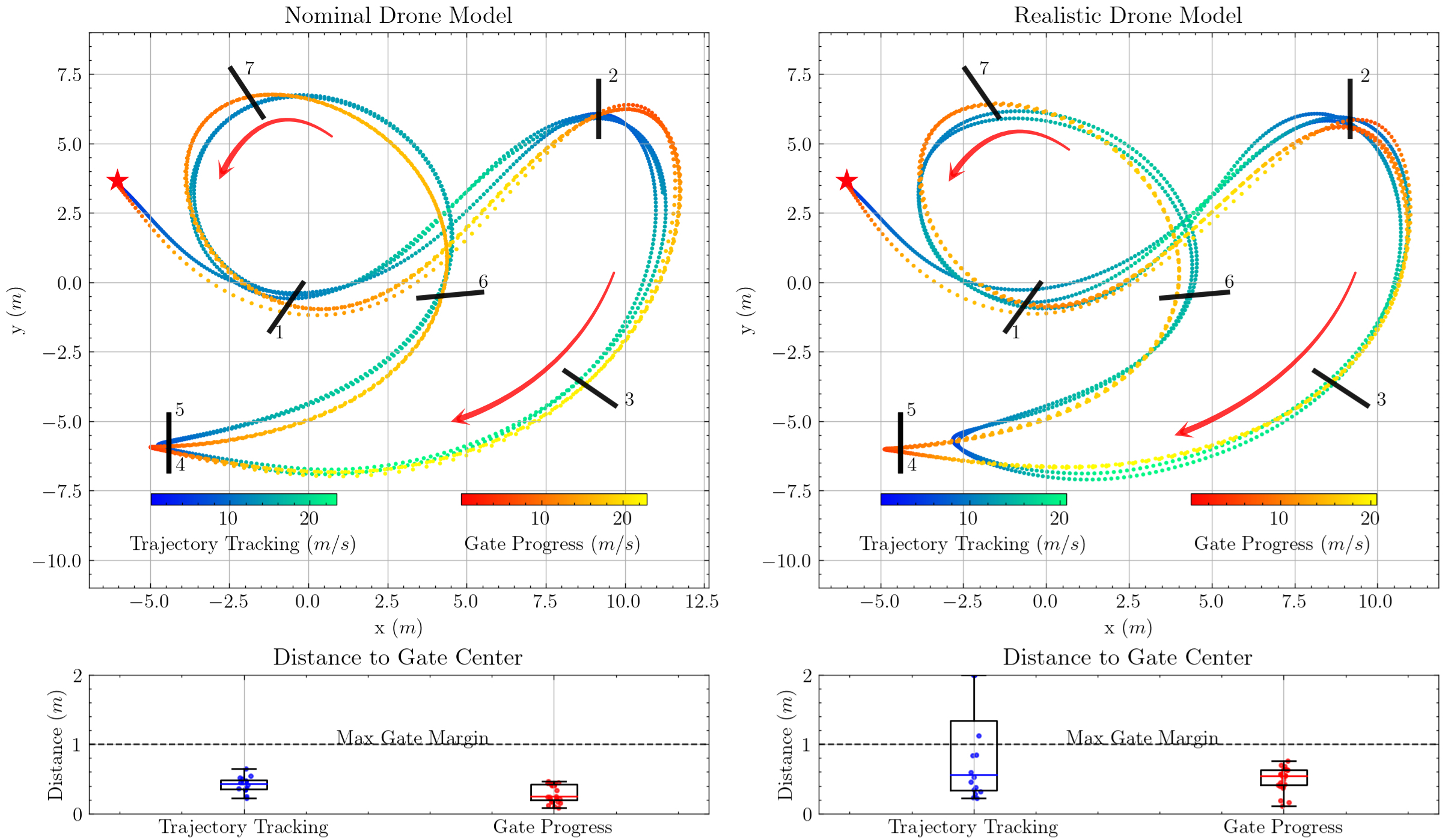}
            \end{subfigure} 
        \end{tabular}      
    }
    \end{tabular}
    \begin{tabular}{p{1\linewidth}}
    \adjustbox{max width=1\linewidth}{
    \begin{tabular}{ c c c | c c  | c c | c c}
            \toprule
            \multirow{3}{*}{Drone Model} 
            & \multicolumn{4}{c|}{ \textbf{Optimization Method}}
            & \multicolumn{4}{c}{ \textbf{Optimization Objective} } \\ \cline{2-9}
            & \multicolumn{2}{c|}{ \textit{MPC Trajectory Tracking}} 
            & \multicolumn{2}{c|}{ \textit{RL Trajectory Tracking}} 
            & \multicolumn{2}{c}{ \textit{RL Gate Progress} } 
            & \multicolumn{2}{c}{ \textit{RL Trajectory Tracking} } \\ \cline{2-9}
            & Lap Time~[\SI{}{\second}] & Tracking Loss
            & Lap Time~[\SI{}{\second}] & Tracking Loss 
            & Lap Time~[\SI{}{\second}] & \# Failed Gates 
            & Lap Time~[\SI{}{\second}] & \# Failed Gates  \\
            \hline
            Nominal & 4.90 $\pm$ 0.08 & 3.32 $\pm$ 3.14 & 4.90 $\pm$ 0.08 & 12.62 $\pm$ 12.69  & 4.99 $\pm$ 0.01 & 0 & 4.90 $\pm$ 0.08  & 0 \\
            Realistic & -- & 71.26 $\pm$ 111.80 & -- & 313.08 $\pm$ 344.88 & 5.07 $\pm$ 0.01 & 0 & -- & 3  \\
            \bottomrule
        \end{tabular} 
        }
    \end{tabular} 
    };
    \node at ([xshift=0.0cm, yshift=-0.05cm]main.north west) [anchor = north west, inner sep = 0pt, outer sep = 0pt, text width = 0.5cm, align = left] {
      \includegraphics[]{Images/A_letter.pdf}%
      };
    \node at ([xshift=10cm, yshift=-0.05cm]main.north west) [anchor = north west, inner sep = 0pt, outer sep = 0pt, text width = 0.5cm, align = left] {
      \includegraphics[]{Images/B_letter.pdf}%
      };
    \node at ([xshift=0.0cm, yshift=-4.4cm]main.north west) [anchor = north west, inner sep = 0pt, outer sep = 0pt, text width = 0.5cm, align = left] {
      \includegraphics[]{Images/C_letter.pdf}%
      };
\end{tikzpicture}
    \caption{ 
    \textbf{Optimization method versus optimization objective.}
    \textbf{(A)} Optimizing the trajectory tracking objective with MPC and RL. Both controllers achieve the optimal lap time when tested with the nominal dynamics model, and both controllers fail when facing unmodeled dynamics at test time.
    \textbf{(B)} Using RL to optimize the trajectory tracking and gate progress objectives. The same optimizer produces a more robust control policy when maximizing the gate progress objective.
    \textbf{(C)} A comparison of lap time, tracking loss, and the number of failed gates. 
    \rebuttal{The red arrows indicates the flight direction and the star symbol shows the starting position. } 
    } 
    \label{fig: objective_vs_method}
\end{figure*}

The second hypothesis considers the difference in the objective that is being optimized by the two methods.
%
%
RL is capable of optimizing nonlinear, nonconvex, and even nondifferentiable objectives.
%
On the other hand, OC typically requires continuity and even convexity in the objective. 
As a result, OC relies on a series of simplifications that express the task objective in terms of intermediate representations, such as reference trajectories. 
%
%
We thus consider the \final{Optimization Objective Hypothesis}: RL yields more robust policies because RL can optimize the task objective directly,
such that the policy is not constrained by intermediate representations and can express a broader range of control responses. RL is more effective than OC because RL optimizes a better objective.

To test this hypothesis, we use the same RL algorithm to optimize two different objectives.
The first is the trajectory tracking objective used by OC. The second is the gate progress objective used by RL.
Here, both RL controllers are trained without domain randomization. 
In Figure~\ref{fig: objective_vs_method}B, we can see how different flight behaviors result from optimizing different objectives. The gate progress policy, designed to prioritize making progress through gates, generates a trajectory that provides a higher margin to the gate boundaries and directs the flight path more toward the center of the gate. This policy performs well in both nominal and realistic simulators, effectively passing all gates and maintaining low lap times. On the other hand, the trajectory tracking policy, which follows the time-optimal trajectory more closely and displays a more aggressive racing behavior, has a slightly larger distance to the gate center and a smaller margin to the gate edges when passing through the gate. This behavior may be riskier, especially in challenging conditions. Despite performing well in nominal dynamics, the trajectory tracking policy fails when tested with the realistic dynamics model. In contrast, the gate progress policy successfully passes all gates and maintains low lap times in this more challenging scenario.

Hence, the optimization objective strongly affects task performance.
OC is subject to inherent limitations due to the decomposition of the problem into planning and control.
When the planned reference, such as a time-optimal trajectory, fails to account for model mismatch and system delay, the downstream controller may be susceptible to failure. In RL can directly optimize a task-level objective, enabling a wider range of control responses to be discovered. Our experiments suggest that this is particularly important when deploying policies in conditions that differ from those encountered during training and in scenarios where small disturbances can have severe consequences. In these cases, the ability of RL to directly optimize task-level objectives can be a \final{substantial} advantage over traditional optimization methods.

\newpage
\subsection*{Value Functions for Different Optimization Objectives}
\label{sec:stuying}

This section studies the state-value function resulting from optimizing different objectives. 
The state-value function estimates the expected cumulative reward that an agent can obtain starting from a particular state \final{when} following a given policy. A high value of the state-value function indicates that the corresponding state is expected to collect more positive rewards and vice versa. Hence, the value function is used to guide the agent's actions toward states that are desirable. 

\begin{figure*}[h!]
\centering
\begin{tabular}{c}
\begin{tabular}{c|c | c}
\includegraphics[width=0.3\linewidth]{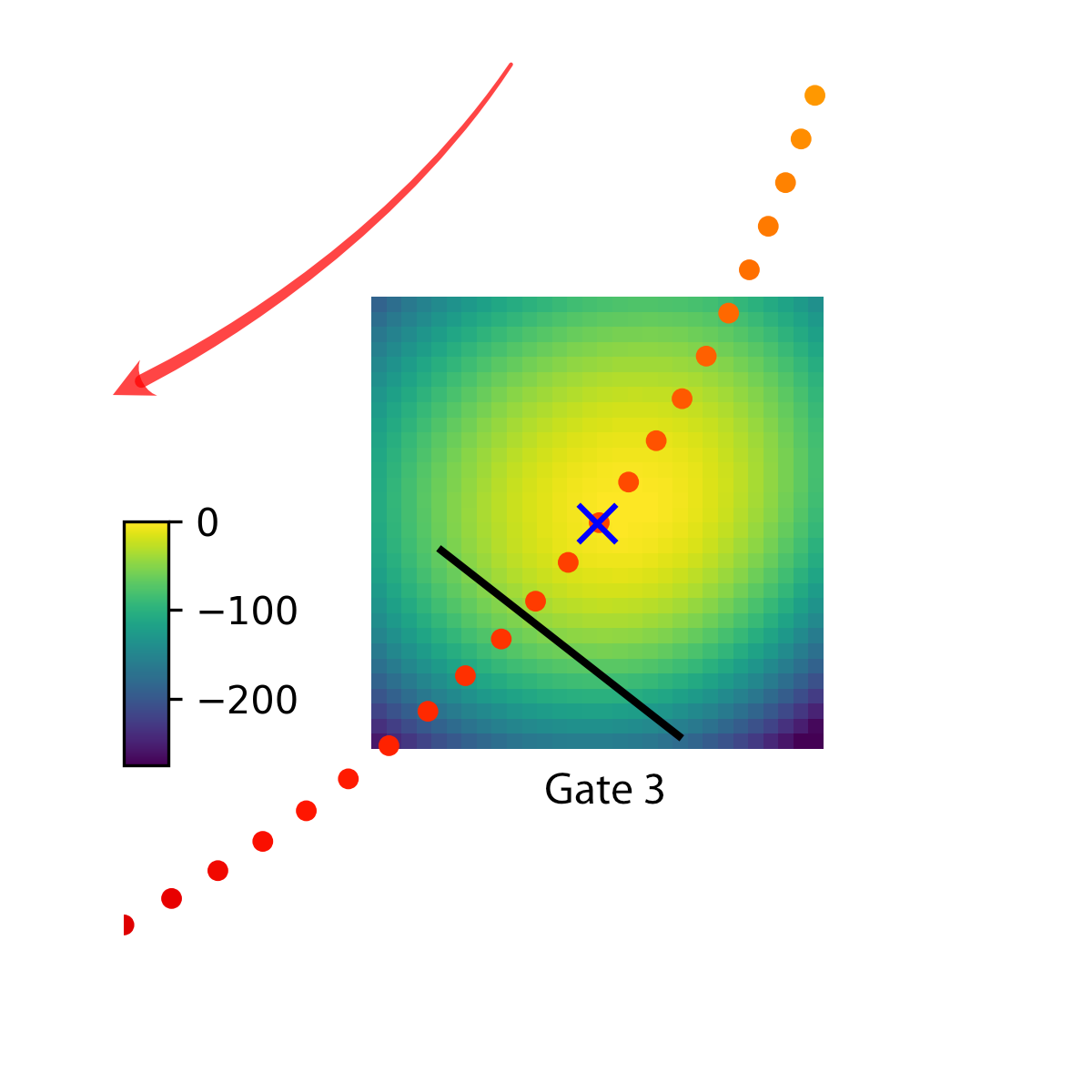} &
\includegraphics[width=0.3\linewidth]{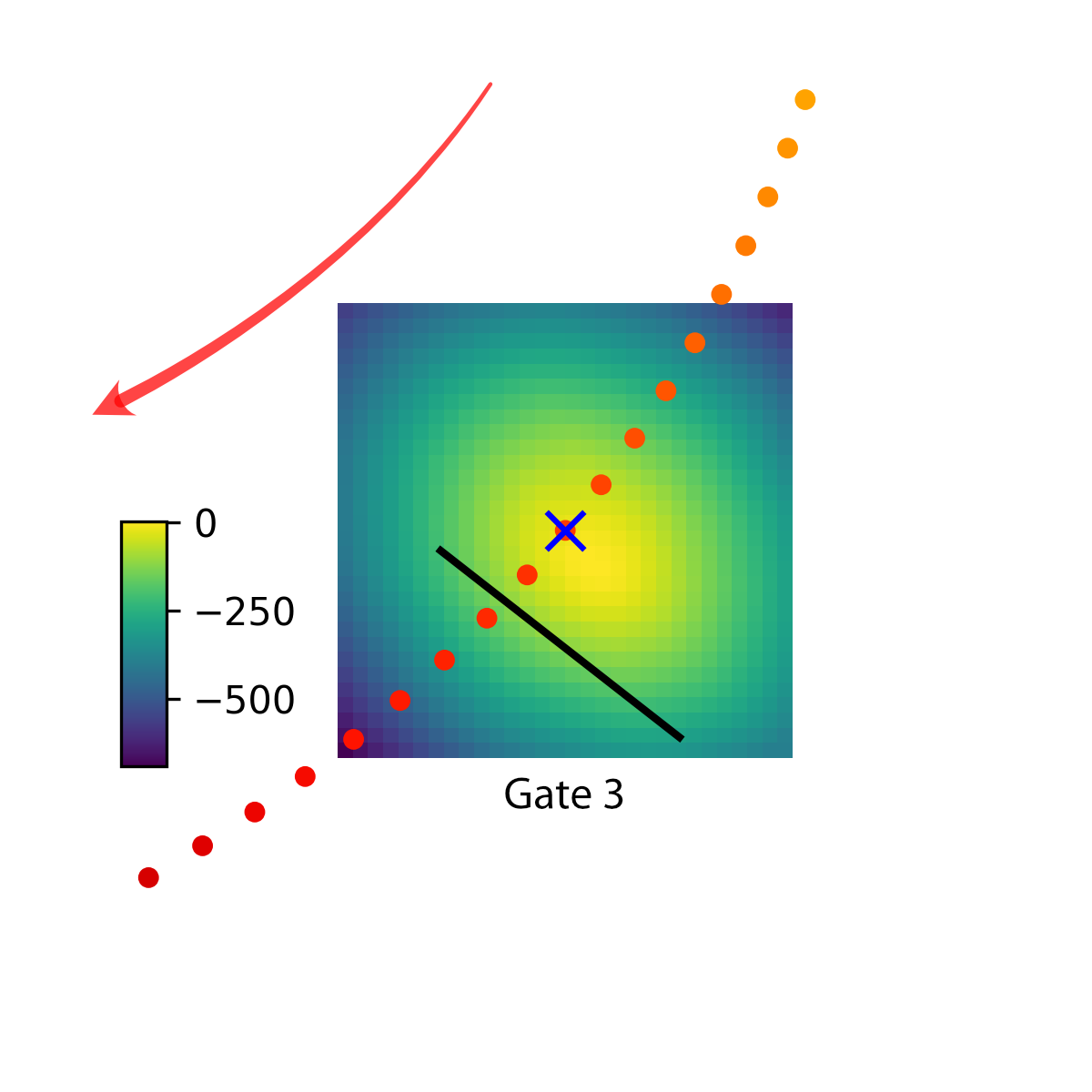} &
\includegraphics[width=0.3\linewidth]{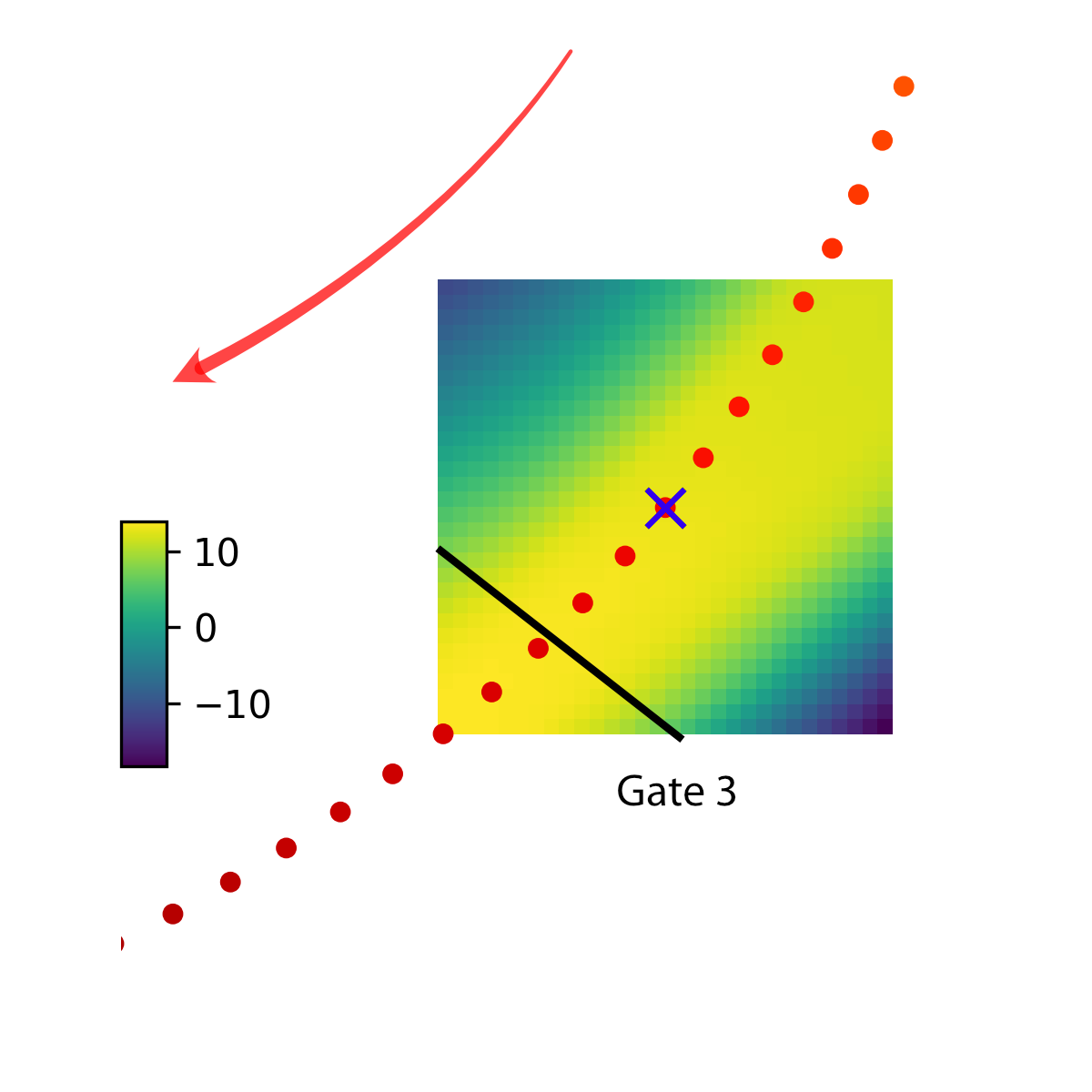}
\end{tabular}
\end{tabular}
\caption{
\textbf{Comparison of the value functions.}
The value function is obtained from optimizing two objectives: Trajectory Tracking and Gate Progress. The Gate Progress objective leads to task-aware behavior, \final{such as maximizing progress through the target gate}, which is not observed in Trajectory Tracking. } 
\label{fig: value}
\end{figure*}

Figure~\ref{fig: value} highlights the differences in the optimized value functions for two different objectives: Trajectory Tracking (with OC and RL, respectively) and Gate Progress (with RL).
We generated synthetic observations by choosing a specific vehicle state (marked as "$\times$") in front of Gate 3 and then sweeping the vehicle's position along the $x$-axis and $y$-axis. 
We then plot the expected value predicted by the critic network in RL or the negative value of the optimal cost in MPC as a distribution of position in the $x-y$ plane. 

As observed, Trajectory Tracking assigns high values when the state is close to the reference state and low values when it is far away, aligning with its objective of minimizing the quadratic loss function between the vehicle state and its reference. Time allocation of the reference state is the incentive for navigating the drone forward, which is done exclusively during the planning stage. However, pre-computed time-optimal trajectories cannot account for model mismatches or variations that are common in real systems. 
The potential mismatch between the time at which the drone is planned to be in a certain state and the actual platform state at that time can lead to the controller tracking unrealistic and possibly infeasible maneuvers, resulting in collisions or cutting corners.

\begin{figure*}[h!]
\centering
\includegraphics[width=1\textwidth]{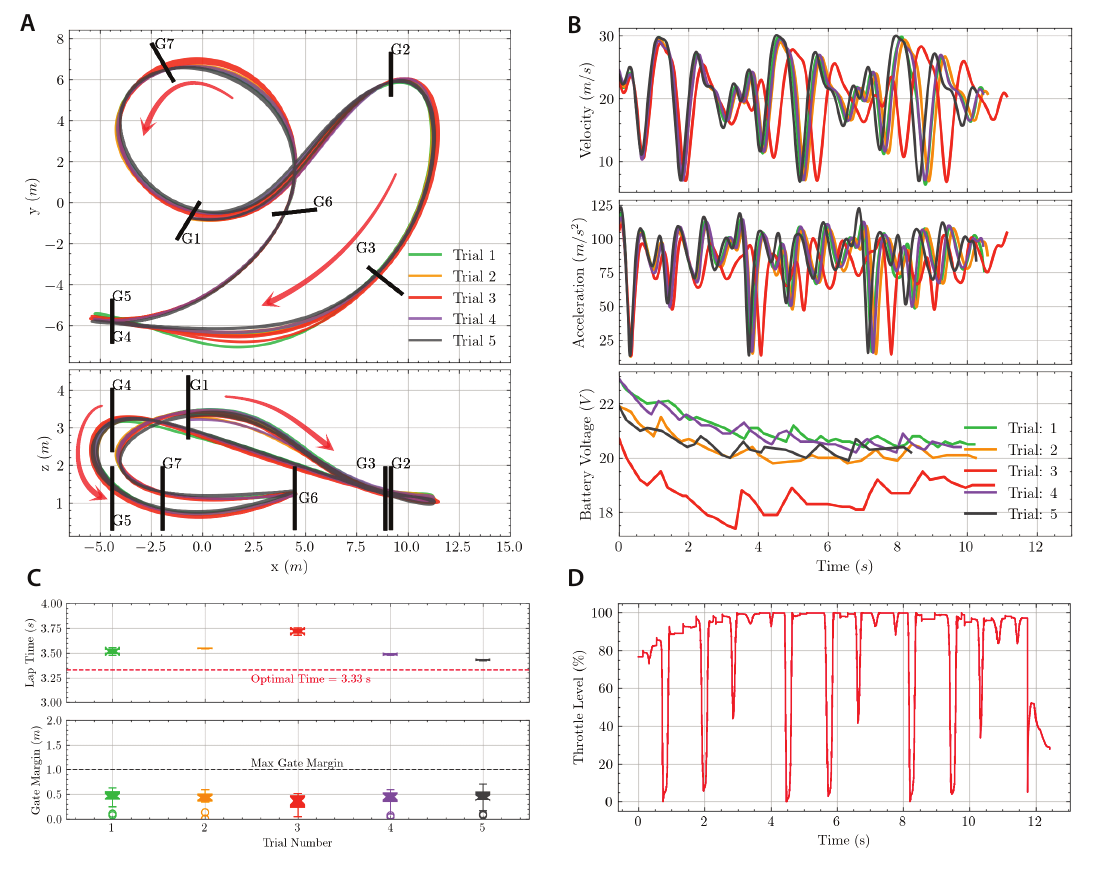}
\caption{ \textbf{Real-world experiments using a high-performance autonomous racing drone controlled by our RL policy.} \textbf{(A)} Flight trajectories projected on the $x-y$ plane (up) and the $y-z$ plane (down). The black lines represent the gates. \textbf{(B)} Velocity, acceleration, and battery voltage over 5 trials. The drone achieves a peak velocity up to~\SI{108}{\kilo\meter\per\hour} and a peak acceleration greater than~\SI{12}{\g}. \textbf{(C)} Lap times and gate passing margins over 5 trials. Despite the battery voltage drop in trial 3, which can be seen in (B), the policy is successful in 100\% of the trials and flies close to the theoretical optimum lap time (dashed red line).
\textbf{(D)} Throttle level in trial 3. The policy pushes the platform to 100$\%$ of the throttle level~-- its physical limit. 
}
\label{fig: realworld_flight}
\end{figure*}

On the other hand, Gate Progress \final{differs from Trajectory Tracking} because it is not limited by a reference trajectory, allowing for a control behavior that emerges directly from learning to optimize the high-level goal. The learned value function in Gate Progress assigns high values to safe and valid states, such as those near the optimal path, and low values to risky states, such as those near the gate border. Unlike Trajectory Tracking, where deviations from the reference state (potentially infeasible) are penalized, Gate Progress allows the vehicle to adapt its behavior freely during deployment. This adaptability leads to more robust performance when facing unexpected disturbances and model mismatches.

\subsection*{Pushing the Limit of Autonomous Racing}
\label{sec:handling}

To push the limit of autonomous racing in the physical world, we have built a racing drone that can produce a maximum thrust-to-weight~(TWR) ratio of 12.
%
This drone has a total weight of only~\SI{0.52}{\kilogram} and can generate a maximum thrust of \SI{63}{N} (given a fully charged battery).
The drone improves upon the highest TWR previously reported~\cite{foehn2022agilicious} by a factor of~2. 
We use this drone to evaluate the characteristics of policies discovered by reinforcement learning.
We use a Vicon motion capture system together with an Extended Kalman Filter to estimate the quadrotor state, including position, velocity, and orientation. 
The RL policy is executed with a feedback control frequency of \SI{100}{\hertz}.
The RL policy is trained in simulation and then transferred to the physical world without fine-tuning.

Note that in order to maintain controllability under model mismatches and disturbances, OC-based systems use a thrust threshold that is lower than what the physical platform can deliver~\cite{Foehn2021science}, or utilize a dedicated real-world controller tuned by a human expert~\cite{Romero2021arxiv}.
In contrast, our RL policy does not artificially constrain the platform or require real-world tuning. 
The RL policy is trained using the full thrust range of the vehicle and then deployed directly on the physical system.  
As a result, it can exploit the vehicle's full performance range.

Figure~\ref{fig: overview} shows two race tracks and a timelapse of maneuvers performed by the RL policy. 
Figure~\ref{fig: realworld_flight} reports experimental results on the Split-S track. 
We fly the policy five times in the physical world; each time, the policy is required to fly three consecutive laps.
In total, the policy flies 15 laps and consistently achieves a $100\%$ success rate.
As shown in Figure~\ref{fig: realworld_flight}D, the vehicle reaches~$100\%$ of the throttle level during high-speed flight -- its maximum performance range. 

Figure~\ref{fig: realworld_flight}B shows the vehicle's velocity, acceleration, and battery voltage during flight. 
The drone achieved a maximum velocity of~\SI{108}{\kilo\meter\per\hour} and a maximum acceleration of \SI{12.58}{g}. 
When flying at such high speed, the battery voltage \final{experiences drops}, resulting in large drops in the produced thrust. 
A particularly \final{large voltage drop} can be observed in trial 3 (shown in Red) due to the use of an old battery in that trial. 
This large voltage drop leads to a longer lap time during this trial.
Nevertheless, the policy maintains control over the vehicle and successfully completes all laps, even in this extreme condition.

Figure~\ref{fig: realworld_flight}C shows the lap times through all trials. For reference, we show the theoretical optimal lap time as a dashed horizontal red line. 
To compute this theoretical optimal lap time, we generate a time-optimal trajectory using an approximate drone model.
Note that the truly time-optimal solution for the physical vehicle is unknown and the theoretical time-optimal trajectory, computed with respect to an approximate model, may not be realizable on the physical vehicle. 
(Attempting to track this trajectory with the physical vehicle using Trajectory Tracking resulted in catastrophic crashes in the real world. Contouring Control with respect to this path could not pass all the gates successfully despite spending a large amount of time tuning the controller on the track by an expert.)

\subsection*{Outracing Human Champions}
\label{sec:outracing}

We have raced the RL policy against three human pilots: Alex Vanover, the 2019-Drone-Racing-League world champion, Thomas Bitmatta, two-time MultiGP-International-Open-World-Cup champion, and Marvin Schaepper, three-time Swiss National champion. The races were held in June 2022 in a flight arena in Zurich. The human pilots were allowed to either use their own drones with similar capabilities to our autonomous platform or drones that are identical to our autonomous platform. 
They were given one week of practice on the track before the races. After this week of practice, each pilot raced against the RL policy in multiple time-trial races.
The race track was designed by Marvin Schaepper; we refer to it as the \final{Marv Track}. 
%
It features a number of challenging maneuvers commonly used in FPV drone racing, including the split-S, power loops, ladders, and flags. 
\begin{figure}[!htp]
\centering
    \begin{tikzpicture}
    \node at (0,0) (main) [text width = 1\linewidth] {
    \fontsize{10pt}{12pt}\selectfont   
    \includegraphics[width=0.85\linewidth]{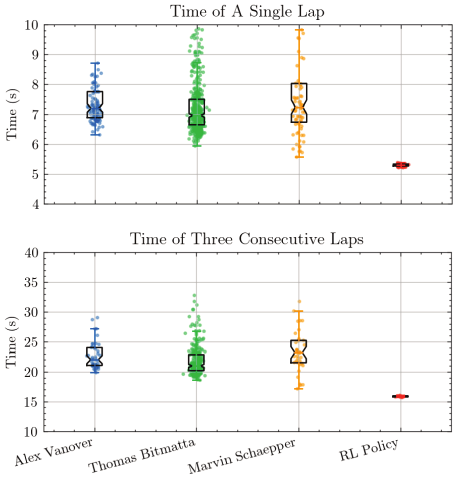}
    \resizebox{\textwidth}{!}{%
    \begin{tabular}{ c | c | c | c }
    \toprule
        Pilot
        & Best Single Lap [\SI{}{\second}]
        & Best Three Consecutive Laps [\SI{}{\second}]
        & Max Linear Velocity~[\SI{}{\kilo\meter\per\hour}]  \\
        \hline
        Alex Vanover & 6.32 & 19.90 &  100.51 \\
        \hline
        Thomas Bitmatta & 5.94 & 18.67 &  102.82 \\
        \hline
        Marvin Schaepper & 5.57 & 17.21 &  105.44 \\
        \hline
        \textbf{RL policy}  & \textbf{5.12} & \textbf{15.59}  &  \textbf{108.72} \\
    \bottomrule
    \end{tabular}
    }
    };
    \end{tikzpicture}
\caption{ \textbf{Comparison of the lap time performance between our RL policy and three professional human pilots. }
Our RL policy outperforms three human pilots in the physical world, albeit with the aid of motion capture and lower latency.}
\label{tab: rl_vs_human}
\end{figure}

\begin{figure}[!t]
\centering
\includegraphics[width=0.5\textwidth]{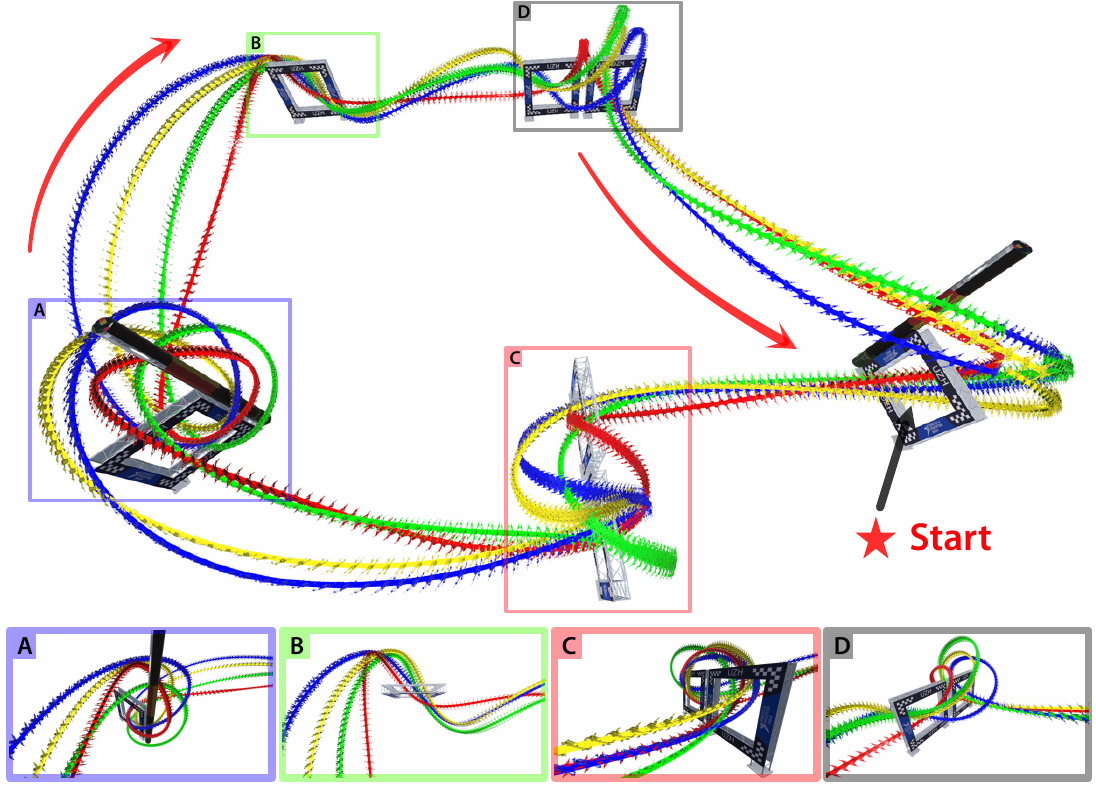}
    \caption{ \textbf{A comparison of the racing trajectories between the RL policy and three human pilots.} Our RL policy outperforms three professional human pilots in the physical world, albeit with the aid of motion capture and lower latency. A visualization of the best trajectories flown by the RL policy ({\color{red}Red}) and the three human pilots: Alex Vanover ({\color{blue} Blue)}, Thomas Bitmatta ({\color{green} Green}), and Marvin Schaepper ({\color{yellow} Yellow}).
    Specific maneuvers executed along the track: \textbf{(A)} ladder, \textbf{(B)} hairpin, \textbf{(C)} straight-enter power loop, \textbf{(D)} side-enter power loop. \rebuttal{The red arrows indicates the flight direction and the \final{star} symbol shows the starting position.} }
\label{fig:best_traj}
\end{figure}

%
As shown in Figure~\ref{tab: rl_vs_human}, the RL policy outraces all human pilots. 
The visualization shows the lap times from all time trials. 
The RL policy achieves the best lap times, with much tighter dispersion. %
It also reaches higher velocities while racing.
The RL policy benefits from near-perfect state estimation from a motion capture system, and lower latency than the human pilots. 
Note that the RL policy is trained in simulation, in only ten minutes on a standard workstation.
Figure~\ref{fig:best_traj} visualizes the best trajectory flown by the RL policy and the three human \final{pilots}, including the world champions of two international leagues. 
The trajectory flown by the RL policy is visibly different from the trajectories flown by the human \final{pilots}. 
The policy flies a shorter overall path and maintains tighter distances to gate borders than the pilots. 
This can be attributed in part to the highly accurate state estimate provided to the RL policy by the motion capture system, which allows the policy to maintain lower margins.

\section*{Discussion} 
\label{sec:discussion}

This paper examined the application of RL and OC to autonomous drone racing. Our experiments show that RL outperforms state-of-the-art OC in this domain. We have investigated the underlying factors responsible for this performance gap, highlighting two possible causes: differences in the optimization method and differences in the optimization objective. Controlled analysis indicates that the key factor is the optimization objective: RL can handle a broader range of objectives, including formulations that directly express the task goal. Such objectives obviate the need for decomposing the problem into planning and control layers that communicate through an explicit intermediate representation such as a trajectory or a path. Directly optimizing the policy for the task goal allows the policy to represent a broader range of adaptive behaviors that provide robustness to unmodeled dynamics and effects.

\rebuttal{Furthermore, RL is modeled as a Markov Decision Process, where the state transition model is stochastic and is formulated via probability. Domain randomization can be used in RL to leverage such probability formulation by training an agent in a variety of environments that simulate different variations of the dynamic system. By training in randomized environments, the agent becomes more robust and adaptable, making it more effective at handling disturbances and model uncertainty. }

As part of this study, we have developed \final{an agile autonomous racing drone that
can produce a maximum thrust-to-weight ratio of 12.}
\rebuttal{We demonstrated a two-layer neural network policy that can handle the vehicle limit in the real world. 
\final{Notably}, the policy is trained purely in simulation, within minutes of training on a standard workstation, and transferred to the real world zero shot.} 
We have conducted time trial races between the autonomous racing policy discovered by RL and three human pilots, including world champions from two international drone racing leagues. The autonomous racing policy consistently outperformed human pilots. \final{Although} the autonomous policy benefited from a number of structural advantages, such as access to perfect state estimation from a motion capture system, this is \final{a milestone} towards the development of autonomous mobile systems that achieve peak performance in the physical world.

\section*{Materials and Methods}
\label{sec:methods}

\subsection*{Optimization Problem}
We formulate autonomous drone racing as a constrained optimization problem. 
We model the drone as a discrete-time dynamical system with continuous states and control inputs, $x_k \in \mathcal{X}$ and $u_k \in \mathcal{U}$ separately. 
Let $f: \mathcal{X} \times \mathcal{U} \rightarrow \mathcal{X}$ be a time discretized evolution of the system such that $x_{k+1} = x_k + \Delta t \cdot f(x_k, u_k)$. 
The system state $x_k$ corresponds to particular instants in time, $t_k = \Delta t \cdot k$ at $k\in [0, N]$. 
The control input $u_k$ affects the system's evolution at time $t_k$. 
Given an initial state $x_0$, the optimality criterion for autonomous drone racing is the minimization of the total time traveled, $t_N$.
%
Apart from respecting the system dynamics $ f(x_k, u_k)$, the optimization has to satisfy additional constraints, such as passing through the gates in the right order without collisions and limiting the thrust to the range that the quadrotor can physically exert.

The minimum time problem for autonomous drone racing is defined as follows.
\begin{align}
    \min_{\tau} & \quad t_N  \notag \\
    \text{subject to:} &  \notag \\
    x_0 = x_{init} & \quad \text{and} \quad x_{k+1} = x_k + \Delta t \cdot f(x_k, u_k)  \notag \\
    g(x_k,u_k) = 0 & \quad \text{and} \quad  h(x_k,u_k) \leq 0
    \label{eq:min_time_opt}
\end{align}
where $g(x_k, u_k)$ and $h(x_k, u_k)$ contain all equality and inequality constraints.
%
The output of the constrained optimization problem above is a time-optimal trajectory $\tau$, which contains a feasible sequence of states and controls:
\begin{equation}
    \tau = (u_0, \cdots, u_{N-1}, x_0, \cdots, x_N).
\end{equation}

Here, the trajectory time $t_N$ is the only term in the cost function. 
In drone racing, a sequence of gates must be passed in a given order, where the gate centers are considered to be waypoints. 
Since multiple waypoints must be passed, the passing time for each waypoint must be allocated as constraints to specific nodes on the trajectory. 
Such time allocation is unknown ahead of time, which renders the optimization problem difficult to solve. 
Recent work~\cite{Foehn2021science} addresses this problem using a numerical optimization scheme that jointly optimizes the trajectory and waypoint allocation in a given sequence.
To achieve this, they developed a progress measure for each waypoint along the trajectory that signals the completion of a waypoint and introduced a complementary progress constraint that limits completion to proximity around the waypoint. 


Given a deterministic dynamical system that is modeled perfectly, the planned trajectory~$\tau$ can be simply executed open loop.
However, since the actual vehicle dynamics $f(x_k, u_k)$ contain unknown disturbances such as aerodynamic effects, directly executing the computed control sequence ($u_k$) will lead to compounding error and a crash. 
To maintain stability and robustness, a feedback controller counteracts unmodeled dynamics by making the control input $u_k = \pi(x_k)$ a function of the observed state $x_k$.
However, the minimization problem \eqref{eq:min_time_opt} cannot be solved in real time. 
Alternative formulations are required for online real-time control. 
%

\subsection*{Optimization Objectives}
\label{subsec: opt_obj}
There are different ways of approximating the problem \eqref{eq:min_time_opt} that support online solutions.

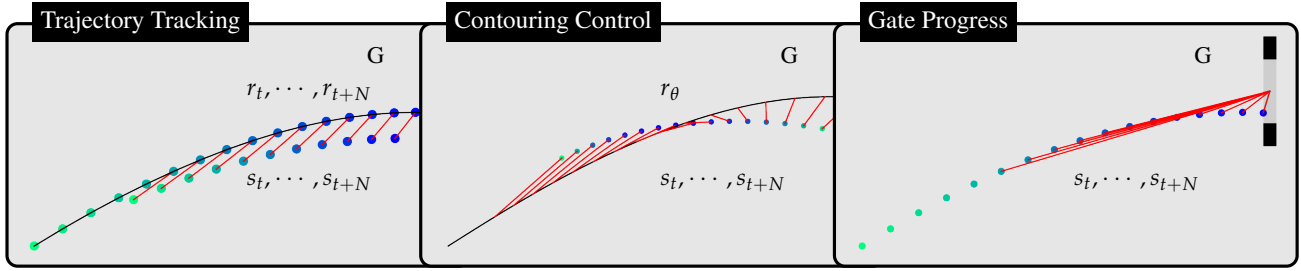
\begin{figure*}[!htp]
\centering
\begin{tikzpicture}

\begin{scope}
    \fontsize{9.5pt}{12pt}\selectfont
    \node [mybox] (box) at (0,0) {%
        \begin{minipage}{0.33\textwidth}
            \includegraphics[width=1\textwidth]{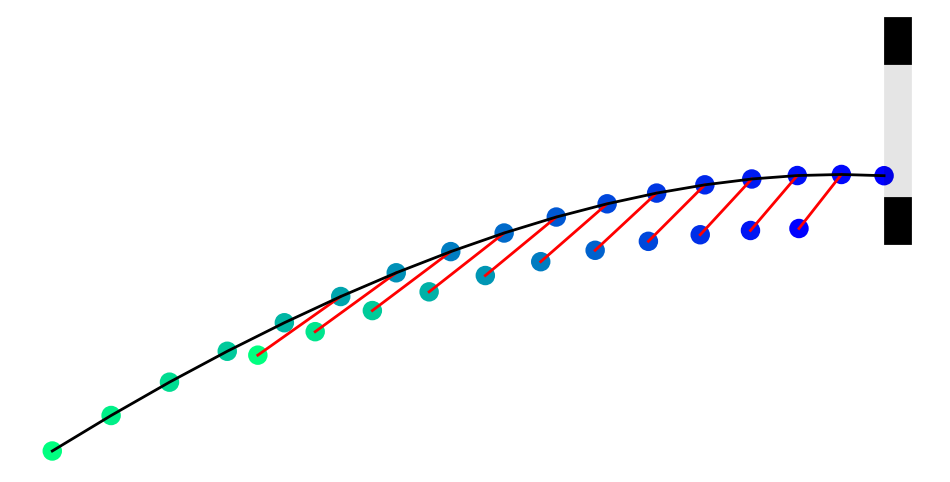} 
        \end{minipage}
    };
    \node[fancytitle, right=10pt] at (box.north west) {Trajectory Tracking};
    \node [mybox] (box) at (5.5,0) {%
        \begin{minipage}{0.33\textwidth}
            \includegraphics[width=1\textwidth]{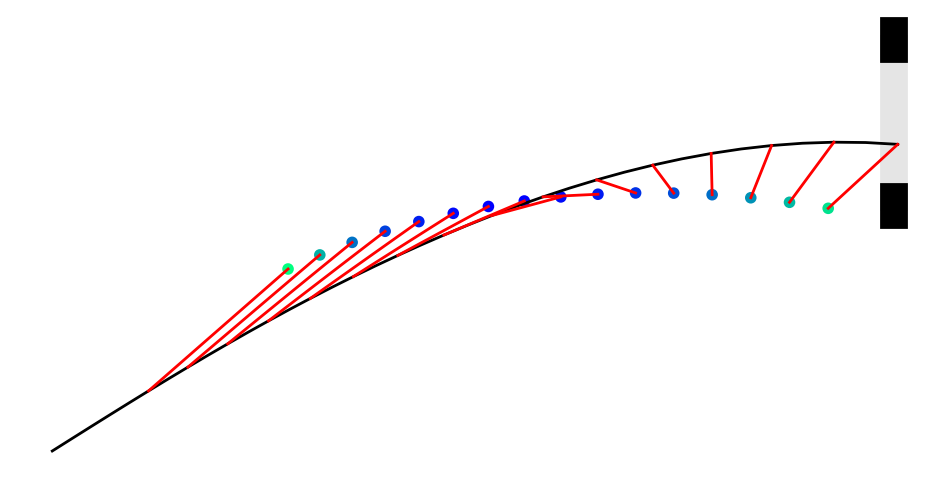}
        \end{minipage}

    };
    \node[fancytitle, right=10pt] at (box.north west) {Contouring Control};
    \node [mybox] (box) at (11,0) {%
        \begin{minipage}{0.33\textwidth}
            \includegraphics[width=1\textwidth]{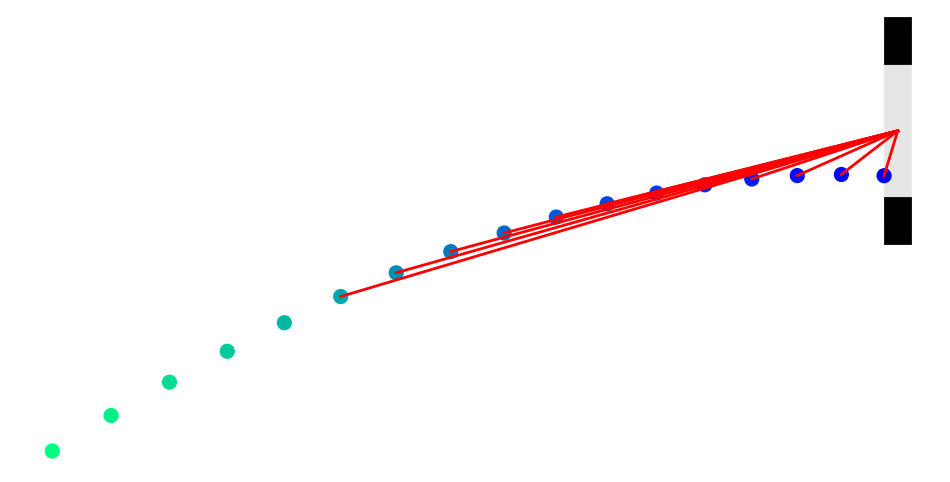}
        \end{minipage}
    };
    \node[fancytitle, right=10pt] at (box.north west) {Gate Progress};
\end{scope}
\begin{scope}[yshift=-1.5cm]
\node [anchor=west] (note) at (1.6, 2.7) {\color{black} G};
    \node [anchor=west] (note) at (0, 1.0) {\color{black} $s_t, \cdots, s_{t+N}$};
    \node [anchor=west] (note) at (0, 2.2) {\color{black} $r_t, \cdots, r_{t+N}$};

    \node [anchor=west] (note) at (7.1, 2.7) {\color{black} G};
    \node [anchor=west] (note) at (5.5, 2.2) {\color{black} $r_{\theta}$};
    \node [anchor=west] (note) at (5.5, 1) {\color{black} $s_t, \cdots, s_{t+N}$};

    \node [anchor=west] (note) at (12.6, 2.7) {\color{black} G};
    \node [anchor=west] (note) at (11, 1) {\color{black} $s_t, \cdots, s_{t+N}$};
\end{scope}

\end{tikzpicture}%

\caption{
\textbf{A visualization of three optimization objectives for autonomous drone racing.} 
All objectives aim to achieve minimum-time flight through the track.} 
\label{fig: racing_objectives}
\end{figure*}

This section discusses three alternative optimization objectives that serve as proxies for the general minimum-time problem.
The first objective relies on first performing a computationally expensive optimization that solves for a time-optimal trajectory and then tracking this trajectory via closed-loop control. 
We refer to this objective as Trajectory Tracking. It uses a quadratic cost to minimize the difference between the vehicle state and the tracked trajectory. 
The second objective is structurally akin to the first, in that a time-optimal trajectory is first optimized offline and then tracked by a closed-loop controller in real time. However, the tracking controller is different, in that the online optimization maximizes the traveled distance along a path while minimizing the deviation from it. Thus the reference is treated as a path rather than a trajectory. This second objective is referred to as Contouring Control.
%
The third objective does not depend on any high-level trajectory or path planner. %
\rebuttal{Instead, it directly maximizes progress toward the next gate center; we call this objective Gate Progress~\eqref{eq: gate_progress}.}
Figure~\ref{fig: racing_objectives} shows a visualization of the three objectives, where $r_t$ represents a time-discretized reference trajectory parameterized by time, $r_\theta$ denotes a continuous path parameterized by $\theta$, $s_t$ are the vehicle states, and $G$ is the target gate.

\subsubsection*{Trajectory Tracking}
Trajectory tracking formulates the autonomous racing problem in terms of tracking a time-optimal reference trajectory. 
This approach divides the task into two stages: planning and control. 
The planning stage maps the task-level objective -- minimizing flight time through the track -- into an intermediate representation in the form of a trajectory.
%
In the second stage, the controller's objective is simplified to a quadratic cost: minimizing the deviation between the vehicle's current state and the reference state along the trajectory.
The objective for the control stage is a sum of quadratic costs within a prediction horizon $N$:
\begin{equation}
    J(x) = \sum_{k = 0}^{N} \Vert x_k - x_{ref,k} \Vert_Q  + \Vert u_k - u_{ref,k} \Vert_R ,
    \label{eq: trajectory_tracking}
\end{equation}
where $x_{ref}$ and $u_{ref}$ refer to the reference trajectory, and $Q$ and $R$ are cost matrices for the state and action costs, respectively. 
Such quadratic costs can be efficiently solved online and are widely used in optimal control for this reason. 
%
%
%

\subsubsection*{Contouring Control}
%
%

Contouring control follows the same basic decomposition of the problem into planning and control. The key difference from trajectory tracking is that the reference is treated as a path rather than a trajectory. The control stage aims to maximize progress along the reference path while minimizing deviation from it. 
In each time step, a new time allocation of the predicted states is computed depending on the current state of the platform.
The objective for the control stage is 
\begin{equation}
    J(x) = \sum_{k = 0}^{N} \Vert \mathbf{e}_k \Vert_Q  +  \| \boldsymbol{\omega}_k \|_R - \rho \theta_N ,
    \label{eq: contouring_control}
\end{equation}
where $\theta_N$ is the total distance traveled along the path within the prediction horizon and $ \mathbf{e}_k = \mathbf{p}_k - \mathbf{p}_{ref}$ is the contouring error, which is the distance between the current position $p_k$ and the reference point $p_{ref}$.
Here, $ \|\boldsymbol{\omega}_k \|_R$ is a penalty on the bodyrate multiplied by a cost matrix $R$. 
%
%
The coefficient $\rho$ trades off between lap time and contouring error. 
%
%

The contouring control objective is less restrictive than the trajectory tracking objective.
It has an additional degree of freedom in selecting the timing for the traversal of the reference path. 
Contouring control can better adapt to model uncertainty and model mismatches than trajectory tracking~\cite{Romero2021arxiv}.
%

\subsubsection*{Gate Progress}
%
The gate progress objective does not rely on a decomposition of the racing problem into planning and control stages. It does not use a reference trajectory or path. Rather, the objective is to directly maximize progress toward the center of the next gate. 
Once the current gate is passed, the target gate switches to the next one.
At each simulation time step $k$, the gate progress objective is defined
%
\begin{equation}
    r(k) = \Vert g_k - p_{k-1} \Vert - \Vert g_k - p_k \Vert - b \| \boldsymbol{\omega}_k \|
    \label{eq: gate_progress}
\end{equation}
where $g_k$ represents the target gate center, and $p_k$ and $p_{k-1}$ are the vehicle positions at the current and previous time steps, respectively. Here, $ b \|\boldsymbol{\omega}_k \|$ is a penalty on the bodyrate multiplied by a coefficient $b=0.01$. 
To discourage collisions with the environment, a penalty ($r(k) = -10.0$) is imposed when the vehicle experiences a collision. The agent is rewarded with a positive reward (($r(k) = +10.0$) upon finishing the race. 

The gate progress objective is nonlinear and nonconvex with respect to the vehicle state. 
It cannot be optimized with classic optimal control methods.
We optimize this objective using reinforcement learning.
\subsection*{Optimization Method}
%
%

This section discusses the two optimization methods we use: model predictive control~(MPC) and policy gradient. 

Model predictive control is based on approximation in value space, where the approximation is done by limiting the number of steps that we look ahead into the future (horizon).
A general formulation for model predictive control is given in \ref{eq:mpc}, where $J(x)$ represents the cost function, $f(x, u)$ is the system equation, $g(x)=0$ is the equality constraint, and $h(x) \leq 0$ is the inequality constraint. 
Given the robot at state $x_0$, a cost function $J(x = x_0)$ is minimized for a finite time horizon $N$ while satisfying a set of system constraints. 
Given the dynamics of the system, MPC outputs a sequence of state and control inputs  $\tau_k = (x_k, u_k, \cdots, x_{k+\Delta t \cdot N})$, which is the predicted trajectory. 
Only the first command is executed on the robot $u^{\ast} = u_t$. 
The robot moves to the next state, and the calculation is repeated again from the new state.

\rebuttal{
MPC involves formulating an analytical and deterministic representation of the system dynamics as $f(x, u)$, followed by online optimization. Unfortunately, exact models are often unattainable in real-world applications due to model uncertainty or external disturbances. To compensate for this, MPC utilizes feedback loop control, runs at high frequencies, and employs conservative assumptions about the robot platform to address modeling errors and counteract potential disturbances.}

In drone racing, the dynamical system and constraints are nonlinear, rendering the MPC problem nonconvex.
Nonconvexity poses challenges for both MPC stability theory and numerical solutions. 
The numerical solution of the MPC problem is typically based on direct optimal control methods using Newton-type optimization schemes, such as direct multiple shooting methods. 
MPC algorithms typically exploit the fact that consecutive optimal control problems are similar to each other. 
This allows for efficient initialization of the Newton-type solver by reusing solutions from the previous iteration, making real-time control feasible~\cite{HPIPM}.
%

Policy gradient approximates a dynamic programming~(DP) problem in the policy space.
A general formulation for policy optimization is shown in \ref{eq:policygradient}.
The goal is to optimize the policy parameters $\theta$ so that the expected return is maximized. In the infinite-horizon case, $R(\tau) = \sum_{t=0}^{\infty} \gamma^{t} r(x_t, u_t)$ defines the objective of the task, where $\gamma \in [0, 1)$ is a discount factor that discounts future rewards.
Here, $p_{\theta}(\tau)$ is the distribution over trajectories $\tau$ collected via policy $\pi_{\theta}$. 
The policy parameters are updated using gradient ascent $\theta = \theta + \alpha \nabla_{\theta}J(\theta)$, where $\alpha$ is the learning rate. 
Unlike online MPC, the optimization in policy gradient is performed offline, ahead of time, such as in simulation. 
After the policy is trained, the computation of the control signal for a given state reduces to a function evaluation: $u^{\ast} = \pi_{\theta}(x_t)$.

\rebuttal{
The main problem in policy gradient is estimating the gradient $\nabla_{\theta}J(\theta)$. 
Using the policy gradient theorem~\cite{sutton2018reinforcement}, the gradient can be computed via 
\begin{equation}
  \nabla_{\theta}J(\theta) = \mathbb{E}_{p_{\theta}(\tau)} \left[ \sum_{t=0}^{T-1} \nabla_{\theta} \log \pi_{\theta} (u_t | x_t) R(\tau) \right].
  \label{eq:pg}
\end{equation} 

Equation~(\ref{eq:pg}) reveals the key insight for policy gradients: the policy update $\nabla_{\theta}J(\theta)$ does not depend on the transition model $p(x_{k+1} | x_k, u_k)$ and can be estimated using sampled trajectories $\tau$. 
Hence, the RL optimization does not require explicit modeling of the system dynamics inside its optimization.
There exist different policy representations for $\pi_{\theta}$.
We use a stochastic policy $\pi \sim \mathcal{N}(\mu_{\omega}, \Sigma)$, where the mean $\mu_{\omega}$ is a function parameterized by $\omega$ and the covariance $\Sigma$ is a vector of parameters.
Given the stochastic representation of the policy and the probability formulation of the state transition model, the policy gradient can readily cope with model uncertainty. }

\paragraph{Model Predictive Control:}
\begin{align}
    & \quad \min_{x(\cdot), u(\cdot)} J(x) = c(x_N) + \sum_{k=0}^{N-1} c(x_k, u_k) \notag \\
    & \textrm{subject to:} \notag \\
    & \quad x_0 = x_{init} \notag \\
    & \quad x_{k+1}  = x_k + \Delta t \cdot f(x_k, u_k)  \notag  \\
    & \quad g(x) = 0 \quad \text{and} \quad  h(x) \leq 0 
    \label{eq:mpc}
\end{align}

\paragraph{Policy Gradient:}
\begin{align}
& \quad \max_{\pi_{\theta}} J (\theta) =  \mathbb{E} \left[R(\tau) | \pi_\theta \right] \notag \\
& \quad  \quad  \quad  \quad  \quad  = \int R(\tau) p_{\theta}(\tau) d \tau \notag \\
& \text{MDP Transition:}  \notag \\
& \quad P_{u}(x_k, x_{k+1}) = p(x_{k+1}| x_k, u_k)  
\label{eq:policygradient}
\end{align}

\subsection*{Optimization Method and Optimization Objective Hypotheses}

At the center of our work is the examination of the fundamental reasons for the success of RL observed in our experiments.
One possible reason is that RL can find better solutions, such as more robust or closer solutions to the optimum, for the same objective.
An alternative reason is that RL can optimize different objectives, whose solutions yield better task performance.

MPC relies on numerical optimization, such as nonlinear programming, where the analytic gradient is used to minimize the cost.  It minimizes a loss function over a short receding horizon. 
On the other hand, policy-gradient-based RL optimizes the objective using stochastic gradient descent, where the policy gradient is estimated via sampled trajectories. 
RL solves the optimization problem offline, ahead of time (in our case, in simulation) and can handle arbitrarily long horizons. 
This leads to our 
Optimization Method Hypothesis: RL outperforms OC because RL, as an optimizer, can achieve better task performance than OC. The difference in the optimization method makes RL more effective than OC.

In addition to the differences in the optimization method, RL and OC typically optimize different objectives.
The family of optimization objectives that can be solved by MPC is limited by the requirements of continuity, smoothness, and convexity.
As a result, MPC is typically coupled with a high-level planner. The controller minimizes a convex objective, typically quadratic, which does not directly express the true task goal. 
%
%
On the other hand, RL can handle a wide range of reward formulations, which need not be convex, continuous, or even differentiable.
%
In the drone racing context, RL need not resort to an explicit decomposition of the problem into planning and control, need not maintain an explicit reference path, and can optimize an objective that directly expresses the task, such as passing each gate as quickly as possible while avoiding collisions.
%
%
We thus arrive at our Optimization Objective Hypothesis: 
RL outperforms OC because RL can optimize a broader range of objectives, including task-level objectives that give the policy more flexibility to discover adaptive control sequences. The optimization objective leads to better task performance.

\subsection*{Reinforcement Learning for Drone Racing}
\begin{figure}[!htp]
\centering
\includegraphics[width=0.5\textwidth]{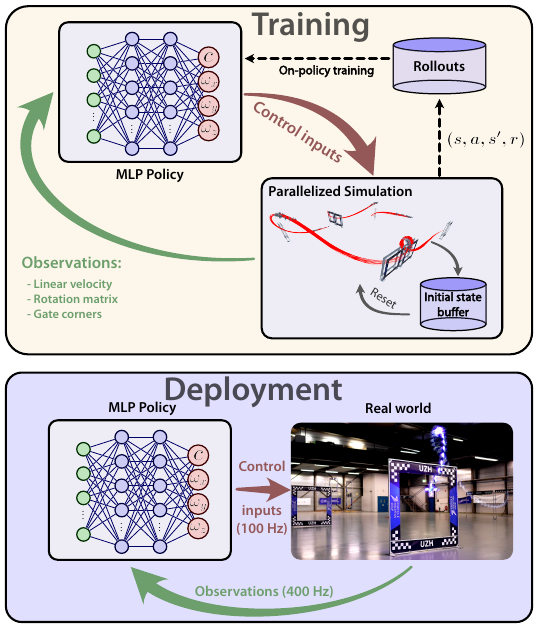}
\caption{ \textbf{An overview of policy training and real-world deployment.} }
\label{fig: policy_training}
\end{figure}

This section provides details about applying model-free RL for optimizing a neural network policy for autonomous drone racing. 
An overview of the approach is given in~Figure~\ref{fig: policy_training}.
First, we model the drone racing problem using the standard Markov Decision Process~(MDP), which is defined by a tuple 
$(\mathcal{S}, \mathcal{A}, \mathcal{P}, r, \rho_0, \gamma)$. 
The RL agent is randomly initialized in a state~$\mathbf{s}_t \in \mathcal{S}$
drawn from an initial state distribution $\rho_0(\mathbf{s})$.
At every time step~$t$, an action $\mathbf{a}_t \in \mathcal{A}$ is sampled from a stochastic policy $\pi(\mathbf{a}_t|\mathbf{o}_t)$ given an observation of the environment~$\mathbf{o}_t$. 
After executing the action, the agent transitions to the next state~$\bs{s}_{t+1}$ with the state transition probability~$\mathcal{P}^{\bs{a}_t}_{\bs{s}_t\bs{s}_{t+1}} = \text{Pr}( \mathbf{s}_{t+1} | \mathbf{s}_t, \mathbf{a}_t)$.
At the same time, the agent receives a reward~${r_t\in\mathbb{R}}$. The goal of RL is to optimize the parameters~$\bm{\theta}$ of a  policy~$\pi_{\bm{\theta}}$
such that the trained policy maximizes the expected return over an infinite horizon. The discrete-time formulation of this objective is 
$ \pi_{\bm{\theta}}^{\ast} = \argmax_{\pi} \mathbb{E}_{\bm{\tau} \sim \pi} \left[ \sum_{t=0}^{\infty} \gamma^{t} r_t \right]$,
where $\gamma \in [0, 1)$ is a discount factor that discounts future rewards.

The main objective of drone racing is to complete a given race track in minimum time, which can be challenging due to the sparsity of rewards when using lap time as the signal. This sparsity makes it difficult to assign credit to individual state-action pairs. 
To address this challenge, we propose using the gate progress reward (Eq.~\ref{eq: gate_progress}).
The gate progress reward incentivizes the agent to fly quickly toward the target gate by maximizing the distance change relative to the gate center. This approach provides a more frequent and informative signal for credit assignments compared to lap time. Additionally, the agent is penalized with a negative reward if it experiences a collision during flight, and it is randomly initialized at a new initial state to encourage exploration. Finally, the agent is rewarded with a positive reward upon successfully completing the maximum episode length. By using the gate progress reward, the agent can learn to navigate the vehicle as fast as possible and avoid collisions.

The observation space consists of two main parts: the vehicle observation~$\mathbf{o}_t^\text{quad}$ and the race track observation~$\mathbf{o}_t^\text{track}$. 
We define the vehicle state as ${\mathbf{o}_{t}^\text{quad} = [\mathbf{v}_{t}, \mathbf{R}_{t}] \in \mathbb{R}^{12}}$, which corresponds to the quadrotor's linear velocity and rotation matrix. 
We define the track observation vector as $\mathbf{o}_{t}^\text{track}=[ \delta \mathbf{p}_1, \cdots,  \delta\mathbf{p}_i , \cdots], \; i \in [1, \cdots, N]$,
where $ \delta\mathbf{p}_i  \in \mathbb{R}^{12} $ denotes the relative position between the vehicle center and the four corners of the next target gate~$i$ or the relative difference in corner distance between two consecutive gates. 
Here $N\in \mathbb{Z}^+$ represents the total number of future gates. 
This formulation of the track observation allows us to incorporate an arbitrary number of future gates into the observation. 
We use $N=2$, meaning we observe the four corners of the next two target gates. 
We normalize the observation by calculating the mean and standard deviation of the input observations at each training iteration.
The agent is trained to directly map the observation to the control, which is expressed as a 4-dimensional vector $\mathbf{a} = [c, \omega_x, \omega_y, \omega_z ] \in \mathbb{R}^4$, representing mass-normalized thrust and bodyrates, in each axis separately. 
We use a two-layer multilayer perceptron (MLP) as the policy network, with 256 nodes in each layer. 
We use a tanh activation function in the last layer of the policy network to produce an output in the range $[-1, 1]$.

\subsection*{Policy Training}
We train the policy using a customized Proximal Policy Optimization (PPO) algorithm~\cite{schulman2017proximal}.
Our customized implementation is based on Stable Baselines3~\cite{stable-baselines3}. 
For training in simulation, we use the open-source Flightmare simulator~\cite{song2020flightmare}. 
Flightmare supports policy rollouts in hundreds of environments in parallel, which significantly speeds up training.
Parallelization is also leveraged to increase the diversity of the sampled trajectories. 

Early in the training, most of the rollouts terminate due to gate and ground collisions.
Consequently, if each quadrotor is initialized at the starting position, the data collection is restricted to a small part of the state space, requiring many update steps until the policy learns to explore the entire track. 
As a countermeasure, we employ an initialization strategy that covers the state space more broadly. 
We randomly initialize the quadrotor with a hovering state around the centers of all path segments, immediately exposing the training to all gate observations. 
Once the policy has learned to pass gates reliably, we store those successful vehicle states in an initial state buffer, which is then used for sampling initial states for resetting the vehicle. 
The use of an initial state buffer greatly improves sample efficiency.

\subsection*{Sim-to-real Transfer}
After training the policy in simulation, we deploy the policy directly in the physical world with no fine-tuning. 
Transferring the policy to the physical world is challenging because the policy operates near the actuator limits of the platform,
%
while the physical vehicle is subject to aerodynamic effects, system delays, and battery voltage drops. 
We make several design choices to facilitate successful sim-to-real transfer. 
First, we identify the physical drone carefully, resulting in a relatively accurate model of the system. 
In addition, we measure the system delay of the real-world drone and simulate the delay during policy training.
Second, the policy outputs control commands in the form of mass-normalized thrust and body rates, which are then tracked by a low-level controller that operates at a higher frequency.
This design enables the use of the same control interface as used by human pilots, while also reducing the sim-to-real gap~\cite{kaufmann2022benchmark}.
Third, we randomize physical parameters that are difficult to identify accurately, including aerodynamic drag and the thrust mapping coefficient. 
Finally, our controller benefits largely from optimizing the gate progress objective, allowing the controller more flexibility to adapt its behavior when facing unknown effects.

\subsection*{Real-world Deployment}
\label{sec:realworld}

We test the policies in a flying arena that spans roughly $30\times30\times8$~\SI{}{\meter} and is equipped with 36 VICON cameras that provide precise pose measurements at a frequency of~\SI{400}{\hertz}.
We use two race tracks, the Split-S track and the Marv
track; both are designed by world-class human FPV pilots (Figure~\ref{fig: overview}). 
Both tracks have 7 gates, but the gates are arranged in different configurations. 
In particular, the Split-S track contains two vertically stacked gates, which require a so-called Split-S manoeuvre.
The Marv track features a number of challenging maneuvers commonly used in FPV drone racing, including the split-S, power loops, ladders, and flags.

We use two different drones: a 4s Drone and a 6s Drone. 
The 4s Drone has a 4-cell battery and is used for the comparisons between RL and OC. 
The 6s Drone has a 6-cell battery and is used for examining the characteristics of the RL policy and for the competition against human pilots. 
Given a fully charged Tattu 6-cell battery, the 6s Drone has a maximum thrust-to-weight ratio (TWR) of 12. 
For comparison, the DJI FPV drone has a maximum TWR of around 4~\cite{foehn2022agilicious}. 
We use the Agilicious control framework for real-world deployment~\cite{foehn2022agilicious}.

\section*{Acknowledgments}

The authors thank Christian Pfeiffer, Florian Trautweiler, Cafer Mertcan Akcay, Thomas L\"angle, and Alex Barden for their contributions to the organization of the race events, the system identification, the real-world deployment, and the drone hardware design. The authors thank Elia Kaufmann for the valuable discussions about the study of optimization objective versus optimization method and Leonard Bauersfeld for his Matlab script for creating Figure~\ref{fig: overview}. Furthermore, the authors thank Yuning Jiang, Benoit Landry, Marco Cusumano Towner, and Drew Hanover for conducting an internal review prior to our initial submission. Finally, the authors thank the human pilots Alex Vanover, Thomas Bitmatta, and Marvin Schaepper for racing against the autonomous drone. 

\textbf{Author Contribution:} Y.S. formulated the main ideas, implemented the system, performed all experiments, analyzed the data, and wrote the paper. A.R. contributed to the conceptualization of the system, the hardware design, optimal control baseline, and paper writing. M.M. contributed to the project conception, revision of the manuscript and provided funding. V.K. contributed to the project conception and direction, high-level design, analysis of experiments, and paper writing. D.S. provided the main idea, contributed to the design and analysis of experiments, to the paper writing, and provided funding. 

\textbf{Funding:} This work was supported in part by the European Research Council (ERC) under grant agreement No. 864042 (AGILEFLIGHT), in part by the European Union’s Horizon 2020 Research and Innovation Program under grant agreement No. 871479 (AERIAL-CORE), in part by the Intel Embodied AI Lab, and in part by the Swiss National Science Foundation (SNSF) through the National Centre of Competence in Research (NCCR) Robotics. 


 \newpage

\bibliography{references}

\end{document}